\documentclass[]{fairmeta}
\usepackage{makecell}
\usepackage{wrapfig}
\usepackage{tabularx}
\usepackage{textcomp}
\usepackage{stfloats}
\usepackage{url}
\usepackage{verbatim}
\usepackage{titlesec}
\usepackage{tocloft}
\usepackage{adjustbox}
\usepackage{multirow}
\usepackage{pifont}
\usepackage{tikz}
\usepackage{comment}
\usepackage{amsmath,amssymb}
\usepackage{colortbl}
\usepackage{natbib}
\usepackage{color}
\usepackage{booktabs} 
\usepackage{hyperref}
\usepackage{graphicx}
\usepackage{subcaption}
\usepackage{multirow}
\usepackage{subcaption}
\RequirePackage{xspace}
\makeatletter
\DeclareRobustCommand\onedot{\futurelet\@let@token\@onedot}
\def\@onedot{\ifx\@let@token.\else.\null\fi\xspace}
\usepackage[most]{tcolorbox}
\usepackage{xcolor}
\usepackage{array}
\usepackage{tabularx}
\usepackage{siunitx} 
\usepackage{makecell}
\usepackage[table]{xcolor}
\usepackage{caption}
\definecolor{headerpurple}{HTML}{d8d2fc}
\definecolor{rowgray}{gray}{0.95}

\def\ie{\emph{i.e}\onedot}

\makeatother

\definecolor{adptorange}{RGB}{248, 205, 172}
\definecolor{cmpblue}{RGB}{189, 215, 238}
\definecolor{cmpblue}{RGB}{189, 215, 238}

\definecolor{our_red}{RGB}{232,157,160}
\definecolor{our_blue}{RGB}{136,206,230}
\definecolor{our_orange}{RGB}{246,200,168}
\definecolor{our_green}{RGB}{178,211,164}

\definecolor{attn_code0}{RGB}{247,215,200}
\definecolor{attn_code1}{RGB}{238,169,139}
\definecolor{mlp_code0}{RGB}{204,201,221}
\definecolor{mlp_code1}{RGB}{102,95,153}
\definecolor{mygray}{HTML}{f0f0f0}

\definecolor{token_blue}{RGB}{84, 120, 140}

\usepackage{pifont}
\usepackage{bbding}
\usepackage{fontawesome}
\usepackage{xspace}

\newcommand{\cmark}{\ding{51}}
\usepackage{float}

\newlength\savewidth\newcommand\shline{\noalign{\global\savewidth\arrayrulewidth \global\arrayrulewidth 1pt}\hline\noalign{\global\arrayrulewidth\savewidth}}

\newcolumntype{x}[1]{>{\centering\arraybackslash}p{#1pt}}
\newcolumntype{y}[1]{>{\raggedright\arraybackslash}p{#1pt}}
\newcolumntype{z}[1]{>{\raggedleft\arraybackslash}p{#1pt}}

\renewcommand{\paragraph}[1]{\vspace{1mm}\noindent\textbf{#1}}
\usepackage{colortbl}
\usepackage{xcolor}
\usepackage{wrapfig}

\renewcommand{\paragraph}[1]{\vspace{1.25mm}\noindent\textbf{#1}}

\usepackage{algorithm}
\usepackage{listings}

\definecolor{codeblue}{rgb}{0.25, 0.5, 0.5}
\definecolor{codekw}{rgb}{0.35, 0.35, 0.75}
\lstdefinestyle{Pytorch}{
    language = Python,
    backgroundcolor = \color{white},
    basicstyle = \fontsize{9pt}{8pt}\selectfont\ttfamily\bfseries,
    columns = fullflexible,
    aboveskip=1pt,
    belowskip=1pt,
    breaklines = true,
    captionpos = b,
    commentstyle = \color{codeblue},
    keywordstyle = \color{codekw},
}

\definecolor{green}{HTML}{009000}
\definecolor{red}{HTML}{ea4335}

\title{Towards Interactive Intelligence for Digital Humans}

\author[1]{Yiyi Cai}
\author[1,2]{Xuangeng Chu}
\author[1]{Xiwei Gao}
\author[1]{Sitong Gong}
\author[1,2]{Yifei Huang}
\author[1,2]{Caixin Kang}
\author[1,2]{Kunhang Li}
\author[1]{Haiyang Liu}
\author[1,2]{Ruicong Liu}
\author[1,4]{Yun Liu}
\author[1]{Dianwen NG}
\author[1, 2]{Zixiong Su}
\author[1,3]{Erwin Wu}
\author[1,2]{Yuhan Wu}
\author[1]{Dingkun Yan}
\author[1]{Tianyu Yan}
\author[1]{Chang Zeng}
\author[1]{Bo Zheng}
\author[1]{You Zhou}

\affiliation[1]{Shanda AI Research Tokyo }
\affiliation[2]{The University of Tokyo }
\affiliation[3]{Institute of Science Tokyo}
\affiliation[4]{National Institute of Informatics}

\teaser{
  \includegraphics[width=\linewidth]{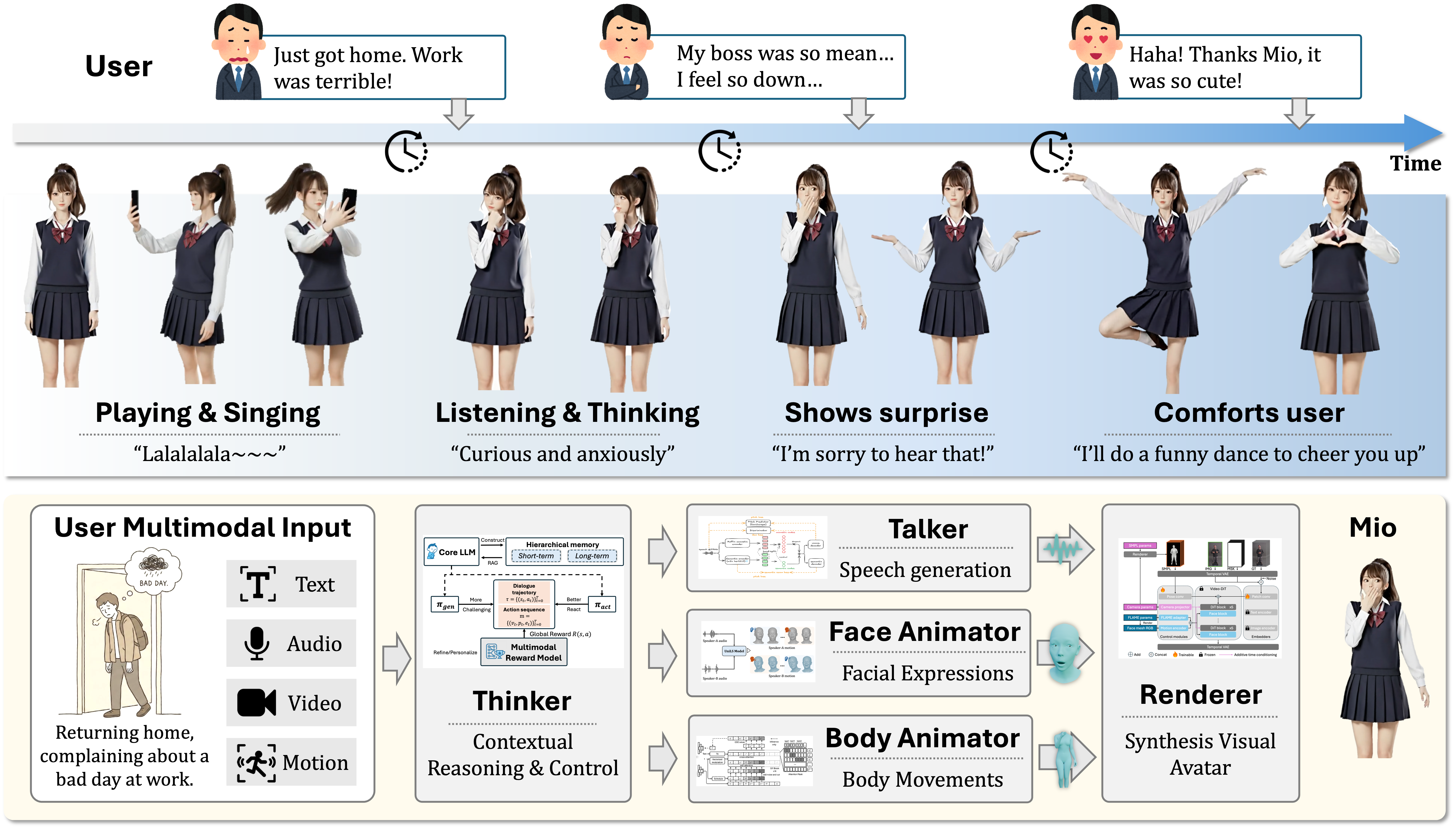}\vspace{5pt}
  \captionof{figure}{\textbf{Towards interactive intelligence with Mio.} The timeline (top) demonstrates a live interaction where Mio adaptively transitions from singing to active listening and comforting in response to user input. The architecture (bottom) features a Thinker that orchestrates multimodal generation via the Talker, Face Animator, Body Animator, and Renderer modules.}

  \label{fig:teaser}
}

\abstract{
We introduce \textit{Interactive Intelligence}, a novel paradigm of digital human that is capable of personality-aligned expression, adaptive interaction, and self-evolution. To realize this, we present \textbf{Mio} (Multimodal Interactive Omni-Avatar), an end-to-end framework composed of five specialized modules: \textbf{Thinker}, \textbf{Talker}, \textbf{Face Animator}, \textbf{Body Animator}, and \textbf{Renderer}. This unified architecture integrates cognitive reasoning with real-time multimodal embodiment to enable fluid, consistent interaction. Furthermore, we establish a new benchmark to rigorously evaluate the capabilities of interactive intelligence. Extensive experiments demonstrate that our framework achieves superior performance compared to state-of-the-art methods across all evaluated dimensions. Together, these contributions move digital humans beyond superficial imitation toward intelligent interaction.
}

\github{\url{https://shandaai.github.io/project_mio_page/}\par\vspace{0.8\baselineskip}}
\date{\today} 

\begin{document}
\thispagestyle{firstheader}
\maketitle
\pagestyle{fancy}

\section{Introduction}

Most existing digital humans remain primarily imitative, reproducing surface patterns of behavior without true understanding of interaction logic. While visual fidelity has greatly improved in recent years~\cite{zhang2024liveportrait,zheng2024genefacepp}, a fundamental gap remains in enabling these avatars to function as responsive, logic-driven entities. To bridge this gap, we introduce \textit{Interactive Intelligence}, a novel paradigm of digital humans that interact seamlessly with users, while possessing personality-aligned expression, adaptive responsiveness, and self-evolution capabilities. This paradigm transforms the digital human from a passive playback system into an embodied agent capable of coherent multimodal engagement within a dynamic narrative context~\cite{park2023generativeagents}.

Current approaches to digital human creation generally fall into two categories: traditional CG pipelines and generative model-based workflows. Traditional CG methods can offer precise control but are hindered by prohibitive production times and reliance on labor-intensive manual processes. On the other hand, workflows utilizing general-purpose multimodal generative models leverage massive audiovisual corpora to accelerate production but remain fundamentally limited to offline generation~\cite{lumiere2024,skyreels2024,yang2024hunyuanvideo,vace,cheng2025wananimate}. Consequently, the resulting characters are primarily imitative rather than autonomous, reproducing surface behavioral patterns without genuine interaction logic. This leaves them incapable of real-time responsiveness and prone to failures in maintaining consistent identity and behavioral coherence over long-term interactions~\cite{tseng2024two,peng2024quantifying}.

Constructing an end-to-end interactive system presents unique challenges across multiple modalities. In response generation, standard LLMs often violate narrative causality (give spoilers) and drift out of persona during extended interactions~\cite{wang2025characterbox,huang2025vinci}. In speech synthesis, existing TTS models lacks efficiently discrete speech representations, hindering the low-latency generation required for fluid conversation~\cite{defossez2022encodec,wang2024valle2,seamless2024, gong2025xy, higgsaudio2025}. In facial animation, a critical issue is the ``zombie-face'' phenomenon, where digital avatars exhibit stiffness and lack natural listening behaviors when not speaking, breaking user immersion~\cite{peng2025dualtalk,agrawal2025seamless,diffposetalk2024}. Furthermore, generating coherent full-body motion remains difficult; autoregressive models often suffer from error accumulation, while standard diffusion models are computationally prohibitive for real-time streaming~\cite{xiao2025motionstreamer,zhang2023t2mgpt,tevet2022mdm,zhang2022motiondiffuse,zhang2025primal}. Finally, rendering these motions into a visual avatar requires maintaining strict multi-view identity consistency, which is often compromised in image-driven diffusion approaches~\cite{Xiong2024MVHumanNet,rudnev2024gsavatar,diffhuman4d2024,paraperas2025arc2face_exp}.

To address these challenges, we introduce Multimodal Interactive Omni-Avatar \textbf{Mio}, a comprehensive framework that models digital humans as autonomous agents with interactive intelligence. We propose a cascading paradigm composed of five specialized modules: \textit{Thinker, Talker, Face Animator, Body Animator, and Renderer}. The \textit{Thinker} serves as the cognitive core, utilizing a hierarchical memory system and diegetic knowledge graph to ensure narrative consistency and personality fidelity. The \textit{Talker} leverages high-fidelity speech representations and produce clear and expressive voice that is well-aligned with the context. The \textit{Face Animator} introduces a unified listening-speaking framework to generate responsive facial dynamics even during silence. The \textit{Body Animator} utilizes a novel streaming diffusion forcing strategy to convert text instructions into physically plausible body motions in real time. Finally, the \textit{Renderer} leverages a parameter-based diffusion transformer to synthesize the visual avatar with precise control over facial and body dynamics while ensuring multi-view consistency.

Extensive quantitative and qualitative experiments demonstrate the superiority of our approach. Our Talker module outperforms exisiting speech tokenizers and auto-regressive TTS models in speech generation metrics with a balanced multilingual capability. The Facial Animator significantly outperforms baselines in listening naturalness, with over 90\% of users preferring our results over existing methods like DualTalk. The Body Animator achieves state-of-the-art motion quality (FID 0.057) on HumanML3D while maintaining the lowest latency and highest smoothness in streaming benchmarks~\cite{guo2022humanml3d}. Furthermore, our Thinker module outperforms general-purpose models like GPT-4o in persona fidelity metrics~\cite{openai2023gpt4}, and the Renderer demonstrates superior multi-view identity preservation compared to recent video diffusion models.

To summarize, Mio represents a fundamental step forward in the evolution of digital humans, harmonizing the often disparate fields of cognitive reasoning and real-time animation. By demonstrating that autonomous agents can possess both narrative depth and physical fluidity, our work paves the way for next-generation applications in virtual companionship, interactive storytelling, and immersive gaming. We believe that Interactive Intelligence will become the defining standard for future avatars, shifting the focus from static appearance to dynamic, meaningful engagement. To support this transition and encourage further exploration within the research community, we make our full codebase, pre-trained models, and the proposed evaluation benchmark publicly available at \url{https://shandaai.github.io/project_mio_page/}.x
\section{Talker} \label{sec:talker}
\begin{figure*}[t]
  \centering
  \includegraphics[width=\textwidth]{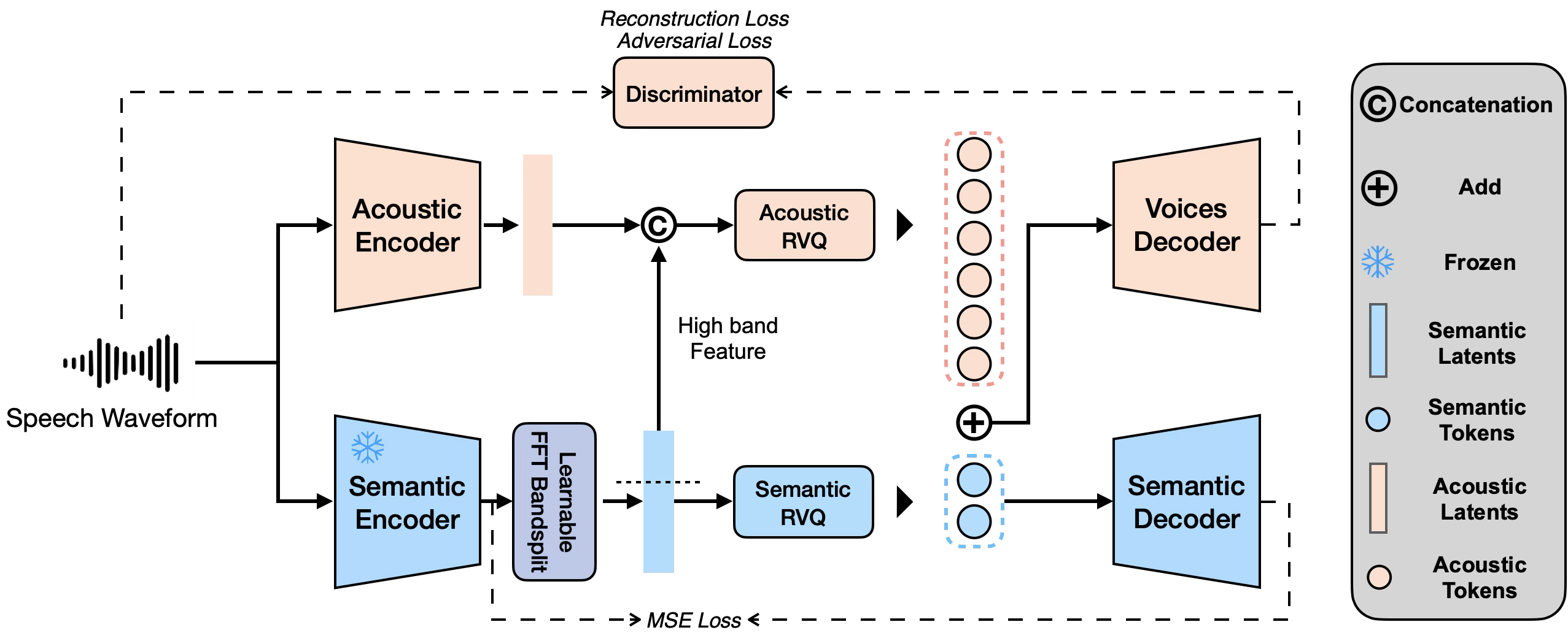}
  \caption{The architecture of Kodama-Tokenizer and training objectives.}
  \label{fig:kodama_codec}
\end{figure*}

The Talker module acts as the speech synthesis engine for Mio. Its primary function is to convert textual output from the Thinker module into natural, high-fidelity speech. Our approach begins by learning efficient discrete speech representations, which are subsequently aligned with semantic content via an auto-regressive (AR) framework. To support real-time, expressive conversational interactions, we prioritize both robust context understanding and generation efficiency.

Crucially, the architecture promotes the disentanglement of semantic and acoustic information through the use of band-splitting and a semantic teacher. This design explicitly separates "what is being said" from "how it sounds," enabling targeted compression for each information type rather than forcing a single representation to resolve both roles simultaneously.

\subsection{Kodama Audio Tokenizer}
\subsubsection{Challenges and Motivation}
Neural audio codecs have recently evolved from tools for efficient waveform compression into the de-facto \emph{audio tokenizers} that connect continuous speech with discrete, latest LLM-ready representations ~\cite{defossez2023high, ye2025codec, ye2025llasa, gong2025xy} all follow a similar recipe: compress speech into low-bitrate token streams that preserve both semantic content and acoustic detail, and then train large language models (LLMs) or speech LLMs to operate directly on these tokens. This paradigm has unlocked highly expressive TTS and multi-modal dialogue systems, but it still faces important trade-offs between compression ratio, reconstruction quality, semantic–acoustic disentanglement, and streaming latency.

Most existing codecs operate at 25--50\,Hz frame rates and $\sim$1\,kbps, which keeps autoregressive sequences relatively long for LLMs. Pushing bitrates lower tends to harm perceptual quality or speaker similarity, while unified encoders and codebooks often entangle semantic and acoustic objectives, making both harder to optimize. In addition, many tokenizers are tuned primarily for offline TTS MOS benchmarks rather than the needs of real-time, dialogue-centric speech LLMs, where shorter sequences, stronger robustness, and easy alignment with text can matter more than absolute reconstruction scores. Therefore, our goal is first to build a speech tokenizer that prioritize low frame rate and bitrate, and by separating semantic and acoustic information, 

\subsubsection{Method}
We propose \textbf{Kodama-Tokenizer}, a neural audio tokenizer designed explicitly for this speech-LLM regime. As shown in~\ref{fig:kodama_codec}, the architecture of the tokenizer module comprises of four components: a semantic encoder, an acoustic encoder, Residual Vector Quantization (RVQ)~\cite{9879532} modules, and a vocoder. The semantic encoder takes 16\,kHz audio waveforms as input, and we borrow the weights from a pretrained W2v-BERT 2.0~\cite{9688253} model to extract SSL features as a semantic teacher. The acoustic encoder is based on the Muffin~\cite{ng2025multiband} codec, which operates on raw waveforms, beginning with a 1D conv that lifts audio into a low-channel representation, followed by four strided Conv1d downsampling stages that progressively compress time while doubling channels up to 512. At each scale, three parallel residual Conv1d blocks with multi-dilation and Snake~\cite{ziyin2020neural} activations refine features, whose outputs are averaged and added back. The final output from acoustic encoder is a 512‑dimensional latent representation at 50Hz. A learnable FFT band-split decomposes the semantic embeddings, where the low-band track is quantized into two RVQ codebooks, while the high-bank track is merged into the acoustic stream. The acoustic feature further uses a 6-codebook RVQ for quantization. Both streams are downsampled to 12.5\,Hz before the RVQ module, with semantic features encoded into two codebooks and acoustic features encoded into six codebooks. We leverage a Vocos~\cite{siuzdak2024vocos} decoder to directly decode the combined embeddings to waveforms, where our main learning objective is to minimize the adversarial loss, derived from the components of the HiFiGAN~\cite{kong2020hifi}, including a multi-period discriminator (MPD) and a multi-scale STFT discriminator (MSD).

Similar to other models ~\cite{kong2020hifi}, we adopt a weighted combination of multiple loss terms as the final loss function formulated by Eq. (\ref{eq:final_dloss}) and Eq. (\ref{eq:final_gloss}) to supervise the training process of our Kodama-Tokenizer. 
\begin{align}
    \mathcal{L}_D & = \mathcal{L}_{adv}(D;G), \label{eq:final_dloss} \\
    \mathcal{L}_G & = \lambda_1 * \mathcal{L}_{adv}(G;D) + \lambda_2 * \mathcal{L}_{mel} + \lambda_3 * \mathcal{L}_{pitch} + \lambda_4 * \mathcal{L}_{fm} + \lambda_5 * \mathcal{L}_{RVQ}, \label{eq:final_gloss}
\end{align}
where $\mathcal{L}_{adv}$, $\mathcal{L}_{mel}$, $\mathcal{L}_{pitch}$, $\mathcal{L}_{fm}$, and $\mathcal{L}_{RVQ}$ denote adversarial loss, mel L2 loss, pitch L2 loss, feature match loss, and quantization loss, respectively. And the corresponding weights of these losses are denoted from $\lambda_1$ to $\lambda_5$. In detail, we adopt the format in LS-GAN \cite{lsgan} to avoid the gradient vanishing for the adversarial training. The formula is shown as
\begin{align}
    \label{eq:adv_loss}
    \mathcal{L}_{adv}(G;D) & = \mathbb{E}_{\boldsymbol{z} \sim \mathcal{N}(0,1)}[(1 - D(G(\boldsymbol{z})))^2], \\
    \mathcal{L}_{adv}(D;G) & = \mathbb{E}_{\boldsymbol{y} \sim p_{data}}[(1 - D(\boldsymbol{y}))^2] + \mathbb{E}_{\boldsymbol{z} \sim \mathcal{N}(0,1)}[D(G(\boldsymbol{z}))^2],
\end{align}
where $G$ and $D$ denote the generator and discriminators, respectively, $\boldsymbol{z}$ is the random noise and $\boldsymbol{y}$ represents the ground-truth speech data. Feature match loss is employed to compel our tokenizer to generate informative discrete tokens, whilst also enabling the reconstruction of high-quality waveforms from these discrete tokens. It can be rerepsented as follows
\begin{align}
    \label{eq:fm_loss}
    \mathcal{L}_{fm} = \mathbb{E}_{\boldsymbol{z}, \boldsymbol{y}}[ & \sum_{i=1}^{L_{MSD}}\frac{1}{N_i}||D^i(\boldsymbol{y})-D^i(G(\boldsymbol{z}))||_1 + \sum_{j}^{L_{MPD}}\frac{1}{N_j}||D^j(\boldsymbol{y})-D^j(G(\boldsymbol{z}))||_1],
\end{align}
where $L_{MSD}$ and $L_{MPD}$ are the number of layers of MSD and MPD, respectively, $D^i(\cdot)$ and $N_i$ represent the feature map and the number of feature map of the $i$-th layer in MSD, and $D^j(\cdot)$ and $N_j$ the feature map and the number of feature map of the $j$-th layer of MPD. 

In order to encourage the model to disentangle semantic and acoustic information as well as focus on generating high-fidelity human speech, semantic reconstruction loss ensures the semantic embeddings alone can perserve the W2v-BERT 2.0 features. Besides, pitch reconstruction L2 loss using the a Crete pitch estimator~\cite{8461329} model to prioritize the reconstruction of prosody and speaker traits. 

Input audio data is resampled to 24\,kHz, therefore our model achieves an extreme $1920\times$ compression ratio by producing tokens at only 12.5\,Hz with 8 codebooks, and its bitrate is as low as 1\,kbps.





\subsection{Kodama TTS}

\subsubsection{Challenges and Motivation}
On top of the tokenizer, we build \textbf{Kodama-TTS}, an LLM-based text-to-speech system that treats Kodama tokens as the speech interface. Similar in spirit to Higgs Audio v2 and MOSS-TTSD, Kodama-TTS models joint text+audio sequences in a unified discrete space and  autoregressively predicts audio tokens, which are then rendered by the decoder of the codec model. Trained on large-scale spoken dialogue data, the 12.5\,Hz token rate is particularly advantageous: it keeps sequences short enough for long-context conversational modeling while remaining expressive enough to capture emotion, speaker identity and language-specific prosody.

Together, Kodama-Codec and Kodama-TTS should form a vertically integrated speech stack. The codec model adopts an extremely compressed, semantically disentangled, streaming-friendly design. Without using a diffusion model that is trained separately to learn the mapping between quantized representation to Mel melspectrograms~\cite{chen2025f5, du2024cosyvoice}, the Kodama-TTS leverages an LLM to fully model the details of generated audio by keeping the original decoder of the codec model. This approach maintains the alignment of the speech encoder and decoder and fills the gap between reconstruction and generation.

\subsubsection{Method}

We use a dual-transformer architecture with a Qwen3-1.7B as the LLM backbone, which predicts the audio token in the first codebook, and then a tiny 200M transformer that also auto-regressively generates the remaining audio tokens conditioned on the backbone's hidden states and previous tokens. The model is adapted to operate over a mixed-modality discrete sequence in which text tokens and Kodama audio tokens share a unified embedding space. This design enables the LLM to directly reason over cross-modal dependencies and to autoregressively produce audio continuations conditioned on linguistic content, prior acoustic context, and conversational history.

\paragraph{Unified Token Embedding and Voice Clone.}
All tokens—whether linguistic or acoustic—are projected into a common hidden space through modality-aware embedding layers. Text embeddings are from the pretraining LLM, while audio-token embeddings are learned from scratch and optimized jointly with the language backbone. At inference time, the model generates audio tokens autoregressively following the textual prompt and optional acoustic exemplars, enabling voice clone ability with in-context learning. Since the LLM directly outputs audio tokens that are native to the integrated Kodama-Codec decoder, the generation path is straightforward and requires no diffusion-based refinement. This reduces latency, preserves reconstruction fidelity, and leverages the codec’s original alignment with natural speech. The model learns to regulate duration, prosody, and expressivity through token-level decisions rather than through external duration models or alignment heuristics.

\paragraph{Fine-tuning}
After the pretraining use full-scale data, we selected expressive speech corpus in our dataset and filtered low quality samples using DNSMOS~\cite{dnsmos} and production quality (PQ) scores from audiobox~\cite{tjandra2025meta}. Low speaker consistency samples that have low intra-utterance speaker similarity, which is calculated using a WavLM-Large model~\cite{wang2021uni} are also removed. To further improve speaker reconstruction and emotion control, speaker embeddings from CAM++~\cite{zheng20233d} and emotion vectors from Emotion2Vec~\cite{xu2018emo2vec} are projected to the embedding space and prepended before the audio tokens.

\paragraph{Learning Objective.}
We formulate the training objective as minimizing the negative log-likelihood of the acoustic tokens. Let $W$ represent the input text sequence, and $A$ represent the target audio sequence of length $T$, where each frame $A_t$ consists of $K=8$ discrete codes from the RVQ layers, denoted as $A_t = \{a_{t,1}, a_{t,2}, \dots, a_{t,K}\}$.

Conditioned on the speaker vector $s$ and emotion vector $e$, the loss function is defined as:$$\mathcal{L}_{\text{TTS}} = - \frac{1}{T} \sum_{t=1}^{T} \sum_{k=1}^{K} \log P(a_{t,k} \mid A_{<t}, W, e, s; \theta)$$

\subsection{Dataset}
We collected open-source corpora and Internet data and curated a dataset of approximately 500k hours. Most data in the dataset has a original sample rate that is higher than 24\,kHz, covering languages including English, Chinese, Japanese, Spanish, German, Spanish, Russian, Korean, Portuguese, French. Genres include audiobooks, podcasts, and content from online video platforms. This large-scale dataset enables both our tokenizer and TTS models to encode and generate diverse and realistic speech signals with a high-level of semantic-prosody alignment.

\section{Facial Animator} \label{sec:facial}
\begin{figure*}[htbp!]
  \centering
  \includegraphics[width=1\textwidth]{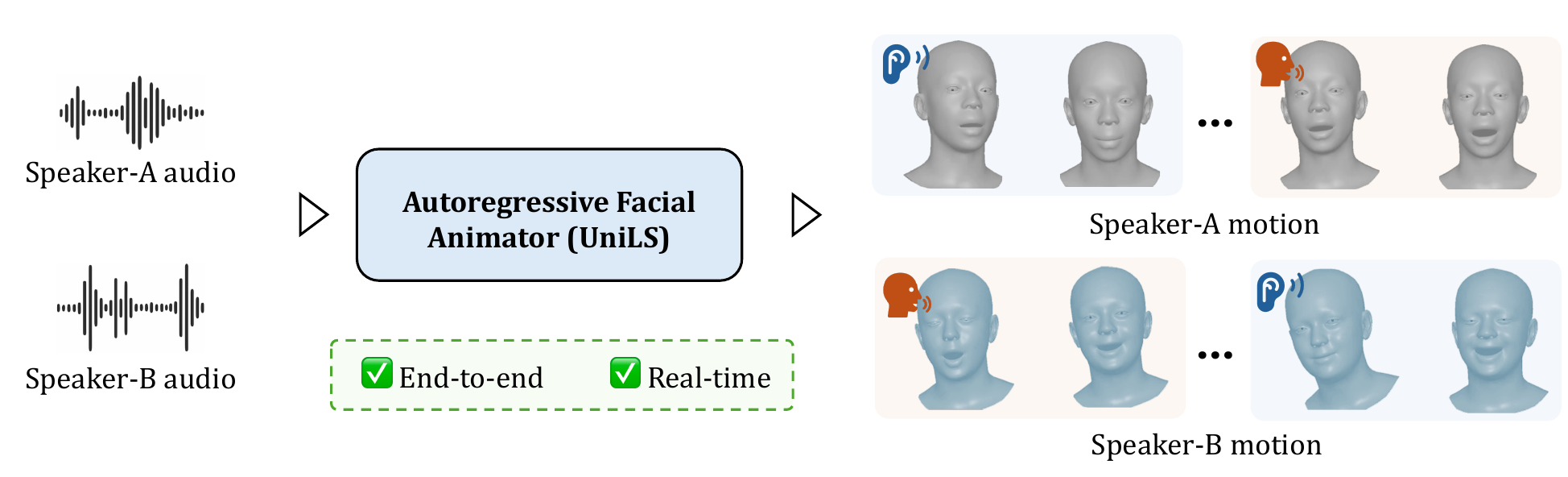}
  \caption{Given dual-track audio inputs from speaker-A and speaker-B, our method (UniLS) autoregressively generates two 3D facial motion sequences. Our method provides an end-to-end framework for unified, real-time speaking and listening motion generation.}
  \label{fig:facial-overview}
\end{figure*}
\subsection{Task Description}
Driven by the audio output from \cref{sec:talker}, we propose the task to generate avatars for unified listening and speaking.
As shown in \cref{fig:facial-overview}, the facial animator aims to generate 3D facial motion sequences for two speakers engaged in a dyadic conversation, driven solely by their respective audio streams.
Given dual-track audio inputs $\mathbf{a}^A$ and $\mathbf{a}^B$, the animator produces two corresponding facial motion sequences $M^A$ and $M^B$, where each sequence contains FLAME-based expression parameters, head pose, jaw pose, and eye-gaze dynamics.

Formally, for each motion chunk of time length $t$, the animator $\mathcal{G}$ autoregressively predicts the next motion chunk for both speakers: 

\begin{equation} \label{eq:audio-gen}
\begin{aligned}
\hat{M}_{t:2t}^A=\mathcal{G}(M_{1:t}^A, \mathbf{a}^A_{1:t}, \mathbf{a}^B_{1:t}, \mathbf{s}^A),\\
\hat{M}_{t:2t}^B=\mathcal{G}(M_{1:t}^B, \mathbf{a}^B_{1:t}, \mathbf{a}^A_{1:t}, \mathbf{s}^B),
\end{aligned}
\end{equation}

where $\mathbf{s}^A$ and $\mathbf{s}^B$ denote style embeddings representing identity-specific motion characteristics \cite{artalk2025}.
The generated motions satisfy two complementary objectives:

\begin{itemize}
    \item \textbf{Speaking behavior generation.} When a speaker is talking, the predicted facial motion should align with the speaker’s own audio, \ie, capturing phoneme–lip correspondence and coordinated head–jaw movements.
    \item \textbf{Listening behavior generation.} When a speaker is not talking, the facial animator should produce natural listening behaviors such as blinks, micro-expressions, subtle head movements, and gaze adjustments. These behaviors should reflect intrinsic motion patterns while being modulated by the other speaker’s audio that provides conversational context.
\end{itemize}

The task therefore requires a unified framework capable of jointly modeling both behaviors, producing continuous, expressive, and realistic 3D facial motions for both sides of the conversation.


\paragraph{Challenge.}
A central challenge in unified listen–speak facial motion generation is the phenomenon of \textbf{listening stiffness}. When an animator is trained end-to-end to map dual-track audio directly to both speakers’ facial motions, the listening motion often collapses into low-variance, nearly static expressions. Such ``zombie-face'' behavior occurs because the audio–motion correlation is fundamentally unbalanced: a speaker’s own audio provides strong phonetic and prosodic cues to drive speaking motions, whereas the listener’s motion is only weakly related to the speech signal.

\subsection{Approach: UniLS}
We design our method by mirroring the natural process of human listening behavior. Natural listening arises from two components:
1) an internal motion prior that reflects spontaneous behaviors such as blinks, nods, and micro-expressions,
and 2) external audio cues that modulate these intrinsic dynamics in response to conversational context.
We develop a two-stage training framework that separately learns these two components.
In Stage 1, we train an audio-free generator to model the internal dynamics of facial behavior.
In Stage 2, we finetune this generator by conditioning on dual-track audios, allowing external speech signals to modulate the facial expression.

\begin{figure*}[t]
	\centering
	\includegraphics[width=\linewidth]{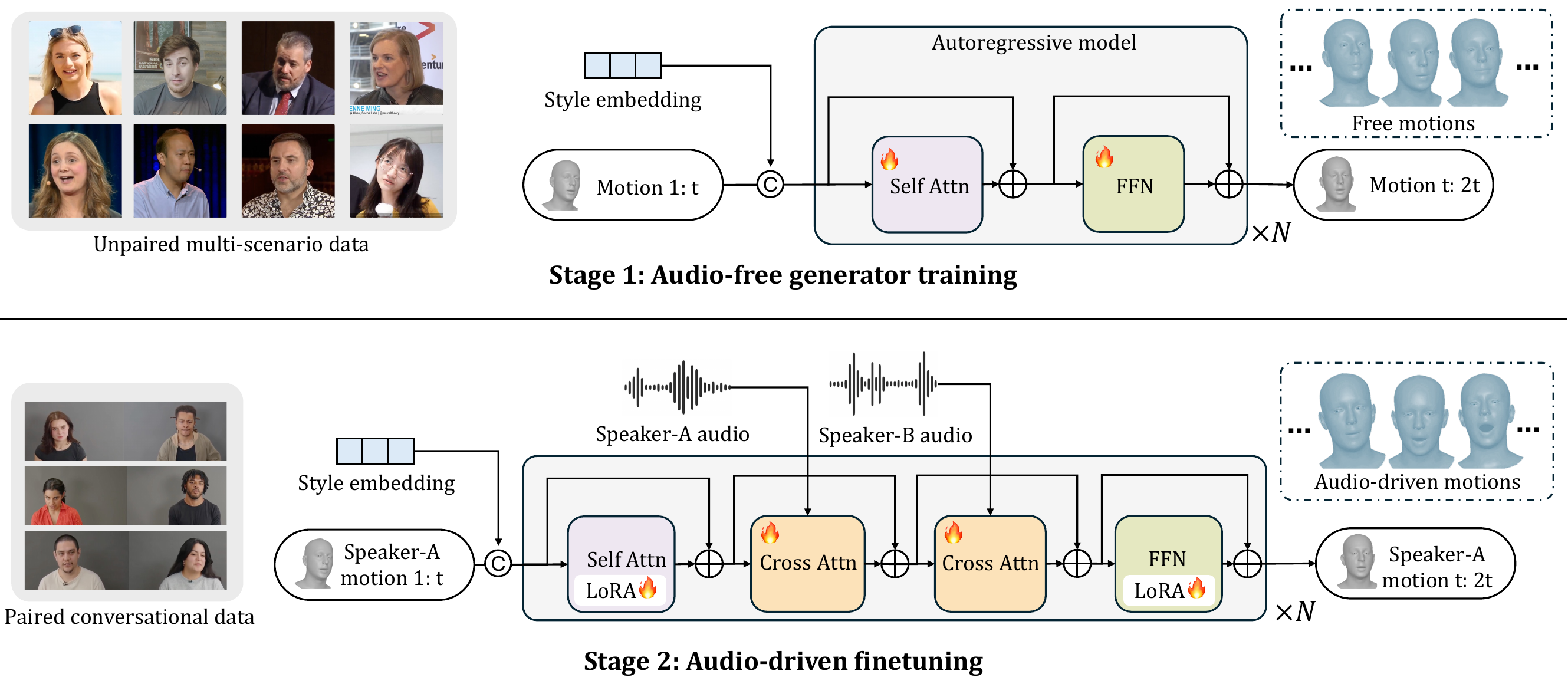}
	\caption{Overview of our two-stage training strategy. Stage 1 trains an autoregressive free generator on unpaired multi-scenario video data without using audio. Given past motions and a style embedding, the model predicts future free motion chunks. Stage 2 finetunes the generator on paired conversational clips by conditioning on speaker-A and speaker-B’s audios through cross-attention, producing audio-driven speak–listen motions.}
	\label{fig:face-method}
\end{figure*}

\paragraph{Stage 1: Audio-Free Generator Training.} In the first stage, we train an audio-free generator to learn internal motion priors that capture natural facial dynamics independent of speech.
This generator is trained on unpaired multi-scenario data, including diverse video sources such as news broadcasts, interviews, streaming content, and casual talking videos, providing diverse facial behaviors across identities and environments.

As shown in \cref{fig:face-method}, the input contains motion chunk $M$ with style embedding $\mathbf{s}$.
This style embedding is learned to encode speaker-specific motion characteristics, following \cite{artalk2025}.
This embedding is concatenated with the input chunk and then fed them into our autoregressive model $\mathcal{G}$.
Our model is transformer-based, composed of stacked self-attention and feed-forward blocks.
At each time step $t$, the generator predicts the next-chunk motion based solely on past motions and the style embedding, \ie:

\begin{equation} \label{eq:free-gen}
\hat{M}_{t:2t}=\mathcal{G}(M_{1:t}, \mathbf{s}).
\end{equation}

\noindent The model is trained with an autoregressive reconstruction loss over each chunk:

\begin{equation} \label{eq:loss}
\mathcal{L}=\sum_{t=1}^T || \hat{M}_{t:2t} - M_{t:2t}||.
\end{equation}

\noindent Through this process, the model learns to produce free motions, which reflects the internal motion prior such as blinking, subtle head movements, and micro-expressions.

\paragraph{Stage 2: Audio-Driven Finetuning.} In the second stage, we finetune the generator to produce audio-driven conversational motions, enabling both speaking and listening behaviors.
This stage uses paired conversational data, where synchronized videos and audios from speaker-A and speaker-B provide the appropriate dynamics required for natural dialogue modeling.

Here we describe the process for generating speaker-A’s motion as an example. As illustrated in \cref{fig:face-method}, the input consists of three components: the motion chunk $M$, style embedding $\mathbf{s}$, and the audios $\mathbf{a}^A, \mathbf{a}^B$ from speaker-A and speaker-B, respectively.
To incorporate audio guidance, we extend the stage 1 architecture by adding two cross-attention layers to each transformer block.
Specifically, one attends to speaker-A’s audio (for speaking behavior) and the other attends to speaker-B’s audio (for listening behavior).
The newly added cross-attention layers are trained from scratch, while the backbone weights inherited from stage 1 are finetuned with LoRA \cite{hu2022lora}.
This design ensures allows the model to adapt efficiently to audio conditioning without overwriting the learned internal motion priors.
Formally, the generating process of stage 2 is expressed as:

\begin{equation} \label{eq:audio-gen}
\hat{M}_{t:2t}=\mathcal{G}(M_{1:t}, \mathbf{a}^A_{1:t}, \mathbf{a}^B_{1:t}, \mathbf{s}).
\end{equation}

\noindent The training objective remains a chunk-wise autoregressive reconstruction loss, which is the same as \cref{eq:loss}.
Generating motions for speaker-B follows the same procedure, with the two audios simply exchanged in their roles.
Through this finetuning, the generator combines internal motion priors learned in stage 1 with dual-track audio guidance, producing smooth and expressive speak–listen motions.

\paragraph{Implementation Details.}
Our facial animator uses a multi-scale VQ-VAE codebook \cite{artalk2025}, which consists of 256 entries, each with a code dimension of 64.
The time window size is 100 frames (4 seconds), and the multi-scale levels are [1, 5, 25, 50, 100].
We used the AdamW optimizer with a learning rate of 1.0e-4 for training the codec, with a total batch size of 64 for 100,000 iterations.
In the two-stage training, we train the autoregressive model using the AdamW optimizer with the same learning rate of 1.0e-4 and a batch size of 128 for 200,000 iterations.
During this training, we employ a frozen wav2vec audio encoder \cite{baevski2020wav2vec}.
All training was conducted on four NVIDIA H200 GPUs, requiring a total of approximately 40 GPU hours (10 GPU hours for the stage 1 and 30 GPU hours for stage 2).

For paired conversational data, we use the Seamless Interaction dataset \citep{agrawal2025seamless}, which offers large-scale dyadic conversational videos.
For multi-scenario data, we additionally adopt four large-scale video datasets: CelebV \cite{yu2023celebv}, TalkingHead-1KH \cite{wang2021one}, TEDTalk \cite{chung2016lip}, and VFHQ \cite{xie2022vfhq}.
To enable 3D facial motion supervision, we apply a carefully designed tracking pipeline to extract per-frame FLAME parameters, including detailed eye-gaze~\cite{huang2018predicting} and head pose annotations~\cite{diffposetalk2024}.
After filtering, we obtain 675.5 hours of conversational data from Seamless Interaction, and 546.5 hours of multi-scenario data from other datasets. The conversational data includes 251.5 hours of speaking motions comprising 22.6M frames, and 406.0 hours of listening motions comprising 36.5M frames. 
For the conversational dataset, we use 622.5 hours for training, 4.8 hours for validation, and 30.2 hours for testing.

\subsection{Result}

\begin{figure}[htbp!]
\begin{center}
\centerline{\includegraphics[width=0.9\linewidth]{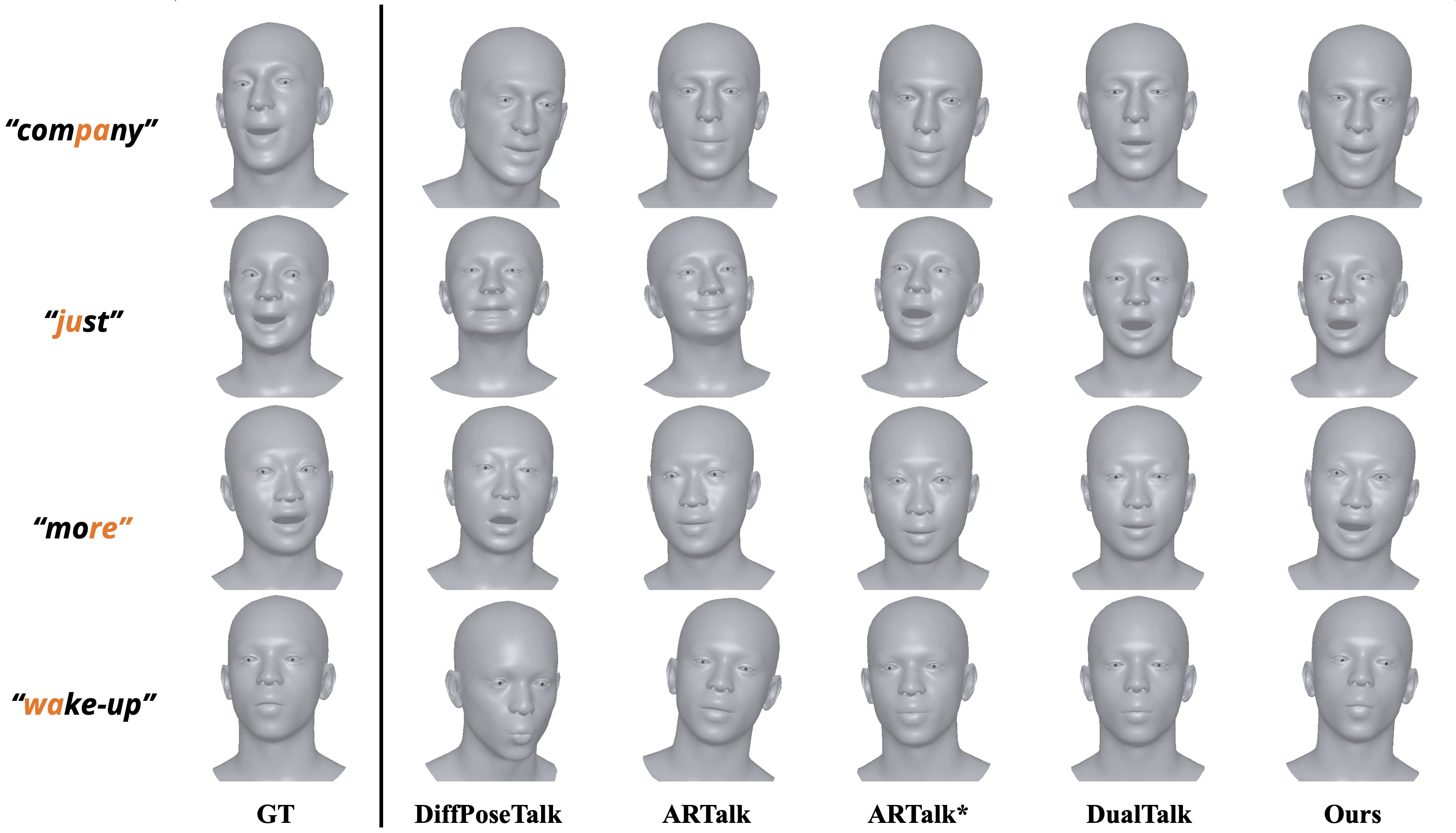}}
\caption{
Qualitative comparison on speaking motions. Our facial animator shows better alignment with the ground truth in expression style and lip synchronization.
}
\label{fig:main_results_speak}
\end{center}
\end{figure}

\begin{minipage}[t]{0.5\textwidth}
\vspace{-22em}
In \cref{fig:main_results_speak,fig:main_results_listen}, we qualitatively compare our method with other baseline methods for listening and speaking motions, respectively.
As shown in \cref{fig:main_results_listen}, we evaluate our facial animator on its ability to generate natural listening motions. ARTalk* \cite{artalk2025} and DualTalk \cite{peng2025dualtalk} exhibit noticeably stiff and low-variance facial behaviors, as highlighted by the red dashed boxes, often remaining close to a neutral expression with limited blinking, head movement, or micro-expressive changes.
In contrast, our animator produces vivid, expressive, and temporally diverse expressions. 
The generated faces exhibit natural head dynamics, mouth shapes, and micro-expressions, demonstrating the effectiveness of our two-stage training in capturing realistic listening behavior.
In \cref{fig:main_results_speak}, our method demonstrates excellent lip synchronization, accurately capturing a wide range of phonetic elements and their associated articulation patterns. Beyond mouth movements, the generated speaking sequences also exhibit realistic facial behavior, such as micro-expressions and head movements.
\end{minipage}
\hfill
\begin{minipage}[t]{.45\textwidth}
\centering
\includegraphics[width=.9\linewidth]{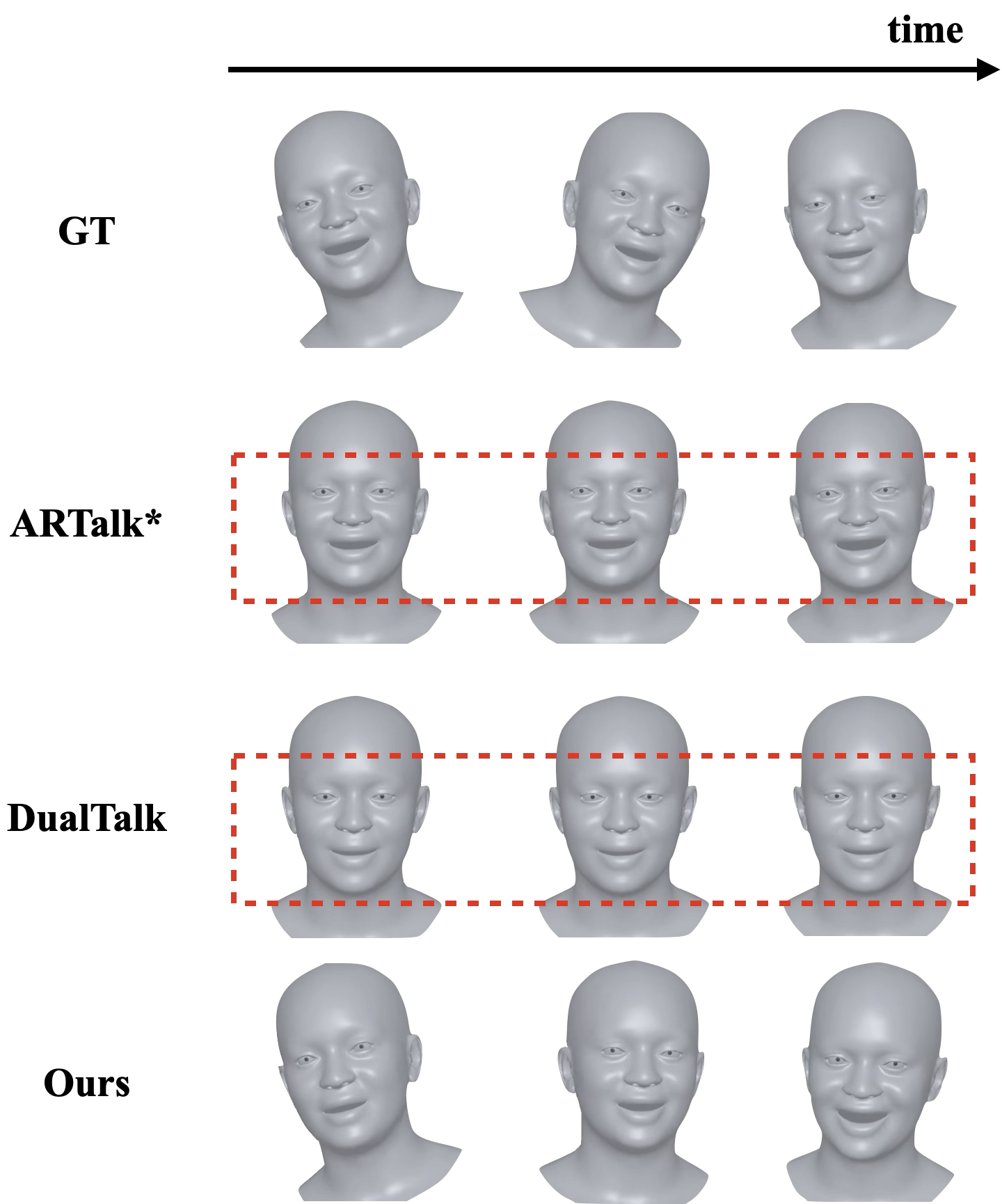}
\captionof{figure}{
Qualitative comparison on listening motions. \textcolor{red}{Red rectangles} highlight motion stiffness over time.
} \label{fig:main_results_listen}
\end{minipage}

\section{Body Animator}

\subsection{Task Description}
\label{sec:body-animator-task}
We formulate \emph{text-controlled, streaming body motion synthesis} as a conditional time-series generation problem. Let $\mathbf{X}^{0:K}$ denote a sequence of body motion states (e.g., joint rotations and positions) and $\mathbf{c}^{0:K}$ denote a time-varying control signal (instruction stream) received from the \textbf{Thinker}. The Body Animator estimates the mapping
$$
\mathbf{X}^{0:K} = g(\mathbf{c}^{0:K})
$$
in a streaming manner, where the output at time $t$ must be generated with strict latency constraints and without access to future control signals. It consumes intent directly from the \textbf{Thinker} (as text prompts or control tokens) and produces physically plausible, seamless full-body motions for the \textbf{Renderer}.

Specifically, we consider the output space of body motion. We adopt the standard motion representation used in HumanML3D, which consists of a 263-dimensional vector including global root velocity, root rotation, local joint rotations, and foot contact information. This high-dimensional continuous space captures the full nuance of human kinematics. The input control signal $\mathbf{c}$ consists of natural language prompts that can be updated at arbitrary time steps $t_k$. The challenge lies in generating $\mathbf{x}_t$ that is coherent with $\mathbf{x}_{<t}$ and aligned with the current active instruction $\mathbf{c}_t$, all while adhering to a frame budget (e.g., 33ms for 30FPS).

To achieve this, we must overcome several challenges:

\paragraph{C1. Real-time streaming under tight latency.}
The generator must operate causally (no future frames), maintain $20$–$60$ Hz output, and amortize model compute so that per-step latency stays below the frame budget. Standard diffusion models require tens or hundreds of denoising steps per frame, which is prohibitive for real-time applications.

\paragraph{C2. Editable control at any time.}
Instructions from the Thinker can arrive mid-gesture (e.g., “walk \textrightarrow{} run \textrightarrow{} wave while turning”). The model must switch goals without visible reset, artifacts, or discontinuities. Unlike offline generation where the text is fixed for the whole clip, streaming requires the model to "steer" the motion trajectory smoothly.

\paragraph{C3. Multi-rate, multi-granularity conditioning.}
High-level intent (style/persona) evolves slowly; action verbs and spatial targets change quickly; prosody and micro-beats from speech can be even faster. Aligning these rates without drift is nontrivial.

\paragraph{C4. Long-horizon coherence.}
Maintaining personality traits and interaction logic across minutes while allowing rapid local edits demands both memory and controllability. A naive frame-by-frame generator often suffers from motion freeze or jitter over long sequences.

\subsection{Approach: FloodDiffusion}
\label{sec:body-animator-approach}

We introduce \textbf{FloodDiffusion}, a framework based on \emph{diffusion forcing} tailored for streaming motion generation. Unlike autoregressive models or chunk-based diffusion which suffer from "first-token" latency or lack of long-term history, FloodDiffusion enables flexible, low-latency generation by allowing different frames to carry different noise levels.

\begin{figure*}[t]
    \centering
    \includegraphics[width=\linewidth]{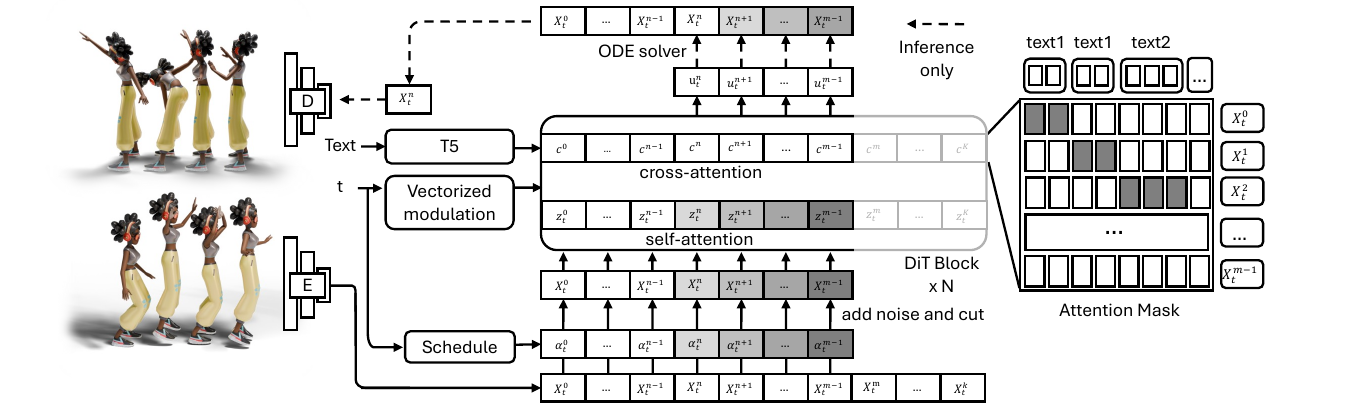}
    \caption{\textbf{Pipeline Overview.} FloodDiffusion encodes the motion stream into a compact latent sequence via a causal VAE. The model predicts velocity for the active window conditioned on context from the Thinker. Key designs include a lower-triangular noise schedule and frame-wise text conditioning. Inference slides the window for streaming output.}
    \label{fig:body-animator-pipeline}
\end{figure*}

\subsubsection{Preliminaries}
We fix the initialization distribution to be standard white Gaussian noise $p_{\text{init}} = \mathcal{N}(\mathbf{0},\mathbf{I})$.
Diffusion models perform distribution matching by transporting $p_0 \sim p_{\text{init}}$ to the data distribution $p_T \sim p_{\text{data}}$ via a time-indexed Gaussian corruption path. For each data point $\mathbf{z} \sim p_{\text{data}}$ and time $t\in[0,T]$, we define
\begin{equation}
p_t(\mathbf{x} \mid \mathbf{z}) = \mathcal{N}\big(\mathbf{x};\, \alpha_t \, \mathbf{z}, \, \beta_t^2 \, \mathbf{I}\big),
\end{equation}
where $\alpha_t$ and $\beta_t$ are scalar schedules. To enable streaming, we extend these to \textbf{Vectorized Time Schedules}. Let $K$ denote the sequence length. We define:
\begin{align}
\bm{\alpha}_t &= [\alpha_t^0,\alpha_t^1,\dots,\alpha_t^{K-1}] \in \mathbb{R}^K \\
\bm{\beta}_t &= [\beta_t^0,\beta_t^1,\dots,\beta_t^{K-1}] \in \mathbb{R}^K
\end{align}
This allows us to assign different noise levels to different frames at the same global "denoising time" $t$.

\subsubsection{Motion Latent Space via Causal VAE}
Instead of discrete tokenization (VQ-VAE) or operating in raw space, we map the high-dimensional motion sequence ($263$D) into a compact continuous latent space ($4$D) using a \textbf{Causal VAE}.
\paragraph{Architecture and Training.} We adapt the causal VAE design from Wan2.1 (video generation) to 1D temporal sequences. The encoder and decoder are strictly causal, meaning the latent $z_t$ and reconstruction $\hat{x}_t$ depend only on $x_{\le t}$.
The training objective is a combination of reconstruction loss and codebook commitment loss:
\begin{equation}
\mathcal{L}_{\text{VAE}} = \| \mathbf{x} - D(z) \|_2^2 + \| \text{sg}[E(\mathbf{x})] - z \|_2^2 + \gamma \| \text{sg}[z] - E(\mathbf{x}) \|_2^2
\end{equation}
where $\text{sg}$ denotes the stop-gradient operator. We use a temporal downsampling factor of 4 and a latent channel dimension of 4. This configuration compresses the 263-dimensional motion data into a highly compact $4 \times T/4$ representation, which significantly reduces the computational burden on the downstream diffusion model while preserving high-frequency motion details. This provides a stable, low-dimensional space for the diffusion model to operate in.

\subsubsection{Tailored Diffusion Forcing}
The core of our approach is a tailored diffusion forcing objective. We discovered that vanilla diffusion forcing (using random schedules) fails for motion data. We instead propose a specific \textbf{Lower-Triangular Schedule}.

\paragraph{Lower-Triangular Schedule.}
Let $n_s$ be the streaming step size parameter. We define the schedule as:
\begin{align}
\alpha_t^k &= \mathrm{clamp}(t - k/n_s, 0, 1) \\
\beta_t^k &= 1 - \alpha_t^k
\end{align}
This schedule creates a "cascading" activation pattern. At any generation step $t$, we can identify three regions:
1. \textbf{Fixed Past ($k < m(t)$)}: Frames are fully denoised ($\alpha=1, \beta=0$).
2. \textbf{Active Window ($m(t) \le k < n(t)$)}: Frames are actively being denoised with varying noise levels.
3. \textbf{Future Noise ($k \ge n(t)$)}: Frames are pure noise ($\alpha=0, \beta=1$).

This structure provides a mathematical guarantee of \textbf{Streaming Locality}. It implies that at time $t$, we only need to compute the update for the active window. Frames before $m(t)$ are finalized and can be sent to the Renderer.

\begin{figure}[t]
    \centering
    \includegraphics[width=0.6\linewidth]{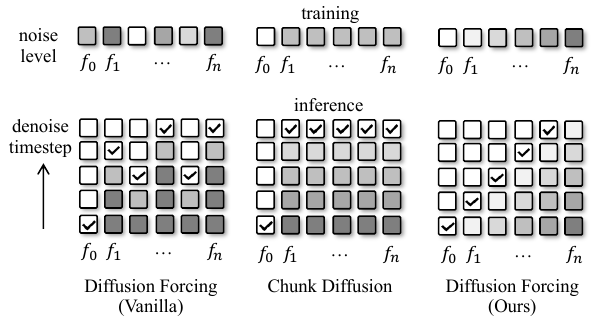}
    \caption{\textbf{Noise Schedule Comparison.} Our triangular schedule (right) denoises only the active window and advances at a constant rate, unlike random schedules or chunk-based diffusion.}
    \label{fig:body-animator-schedule}
\end{figure}

\paragraph{Bi-directional Attention.}
We employ a Diffusion Transformer (DiT) backbone. A critical finding is that within the active window, \textbf{bi-directional attention} is essential. Although the overall system is streaming/causal, the frames \emph{currently being denoised} benefit massively from attending to each other. Restricting attention to be causal within the denoising window degrades performance significantly (FID drops from 0.057 to 3.37).

\subsubsection{Time-Varying Text Conditioning}
To handle changing instructions from the Thinker (e.g., "walk" $\to$ "wave"), we implement a \textbf{frame-wise text conditioning} mechanism.
\begin{itemize}
    \item \textbf{Text Encoding}: We use a T5 encoder to process text prompts into embeddings.
    \item \textbf{Condition Fusion}: We use a biased attention mask where each motion frame $k$ attends to the text prompt active at time $k$.
    \item \textbf{Handling Transitions}: When the Thinker updates the prompt, the new embedding is seamlessly integrated for future frames. The overlapping active window ensures a smooth transition between the old and new motion styles without explicit "prompt refresh" logic.
\end{itemize}

\subsubsection{Streaming Inference Algorithm}
During inference, we slide the active window forward by a fixed step $\Delta t$. This process allows for continuous generation of infinite-length sequences.
\begin{itemize}
    \item \textbf{Step 1: Initialize}. Start with a buffer of pure Gaussian noise.
    \item \textbf{Step 2: Window Identification}. For each simulation step $t$, determine the indices of the active window $[m(t), n(t)]$ using the lower-triangular schedule definitions.
    \item \textbf{Step 3: Predict Velocity}. Run the DiT model. Thanks to Streaming Locality, we only need to compute the output for frames in the active window. The model takes the current noisy latents $\mathbf{z}_t$ and the active text embeddings $\mathbf{c}_t$ as input.
    \item \textbf{Step 4: Denoise Update}. Apply the numerical solver step (e.g., Euler method) to update the latents: $\mathbf{z}_{t+\Delta t} = \mathbf{z}_t + \mathbf{v}_t \cdot \Delta t$.
    \item \textbf{Step 5: Shift and Commit}. Slide the window forward. Frames that exit the window ($k < m(t+\Delta t)$) are considered "committed" and are decoded by the Causal VAE decoder to produce final motion frames for the Renderer.
\end{itemize}
This pipeline yields a constant computational cost per step and strictly bounded latency (approx. $n_s$ frames), avoiding the generation pauses typical of chunk-based methods. The lower-triangular schedule ensures that by the time a frame exits the active window, it has been fully denoised ($\alpha=1$).

\subsection{Results}

\begin{figure*}[t]
    \centering
    \includegraphics[width=\linewidth]{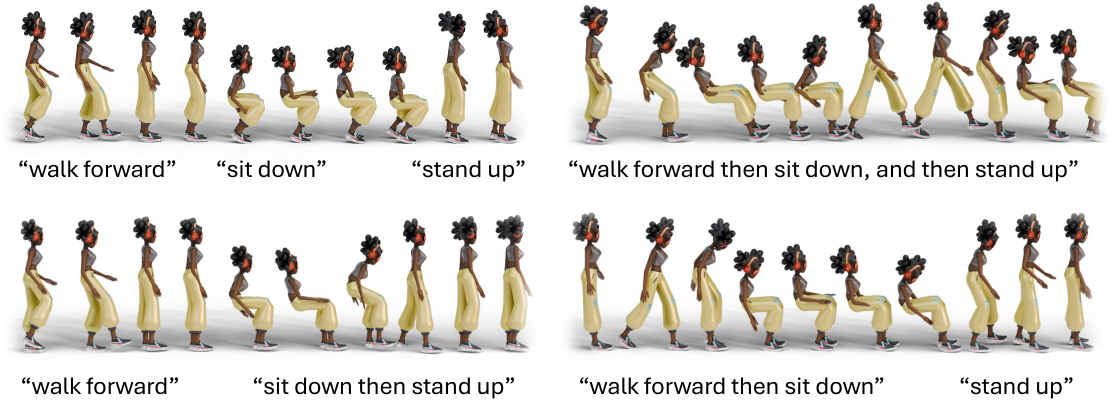}
    \caption{\textbf{Comparison of time-varying conditioning.} Our model generates different resulting motions from the same text prompts based on their delivery timing. (Top Left) Prompts are given separately at different frames. (Top Right) All conditions are fed as a single prompt at once. (Bottom Left) Two separate prompts are input early in the sequence. (Bottom Right) The same two separate prompts are input later in the sequence.}
    \label{fig:body-animator-result1}
\end{figure*}

\begin{figure*}[t]
    \centering
    \includegraphics[width=\linewidth]{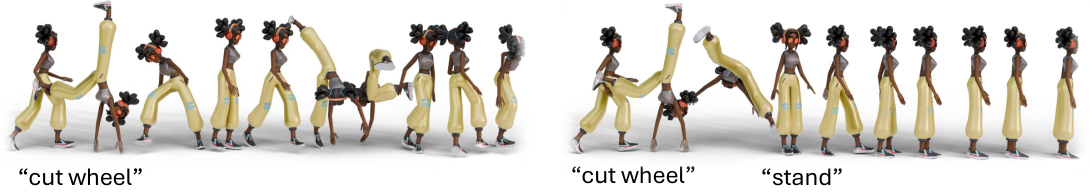}
    \caption{\textbf{Comparison of long sequence generation.} (Left) our model will continue to repeat the motion in text prompt if without new prompts come. (Right) in real application, our model could stop current motion by explicitly giving the rest style prompt, such as ``stand''.}
    \label{fig:body-animator-result2}
\end{figure*}

The Body Animator, powered by FloodDiffusion, converts a live, editable instruction stream into physically plausible, personality-consistent body motion in real time. By combining a causal VAE with a tailored diffusion forcing scheduler, we deliver an industry-first system that supports fully online, instruction-editable body animation with SOTA quality (0.057 FID) and strict frame-rate adherence. This component serves as the robust physical actuator for the broader digital human system.

\section{DiT-based Rendering}
\subsection{Task Description}

\begin{figure}[tbhp]
    \centering
    \includegraphics[width=0.9\linewidth]{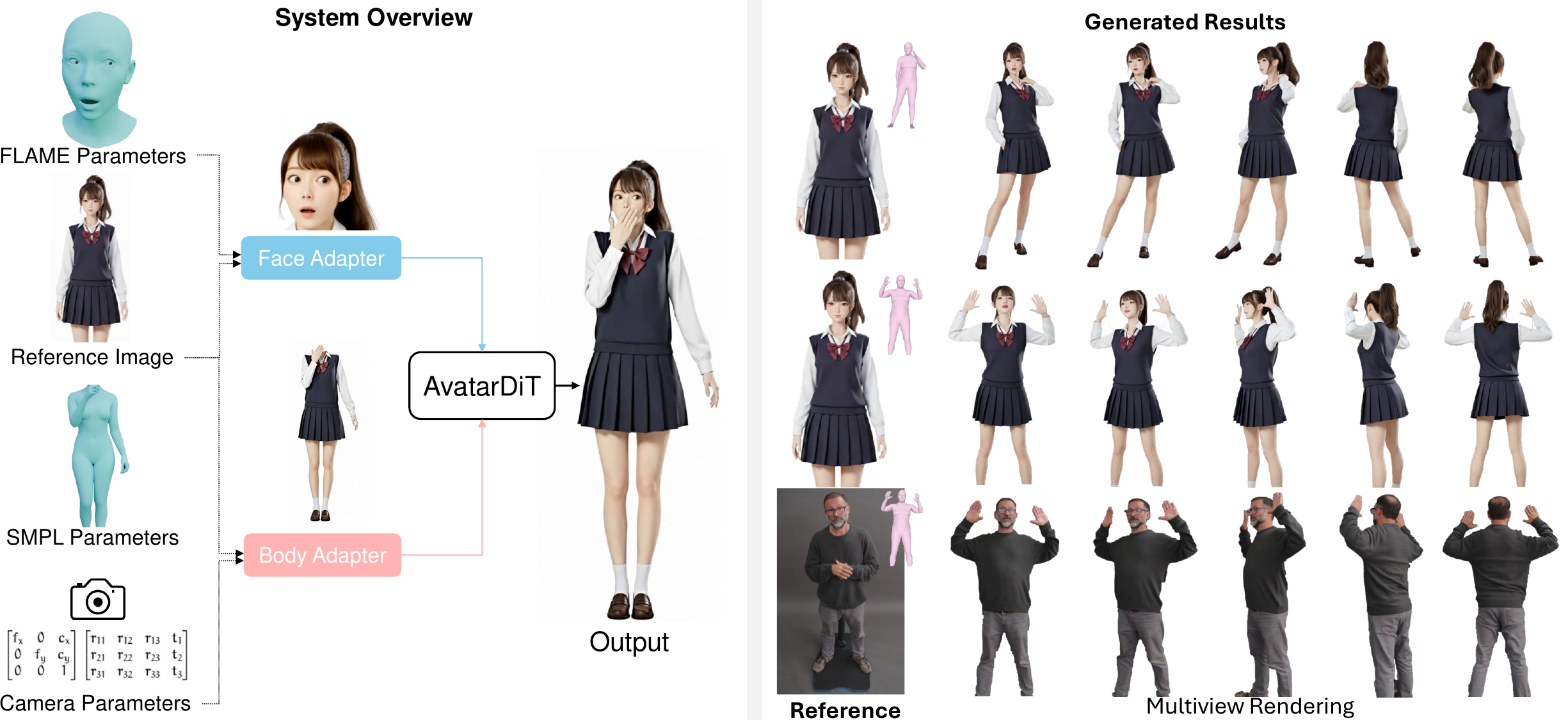}
    \caption {Overview of AvatarDiT and results showcasing multi-view consistency of the proposed framework.} 
    \label{fig:overview}
\end{figure}

The goal of the DiT-based Rendering module is to synthesize high-fidelity, identity-consistent human video frames from a stream of motion parameters provided by the embodiment modules (Facial Animator and Body Animator). Unlike conventional image- or video-driven approaches that rely on reference frames or driving sequences, our renderer operates purely on parameterized 3D control signals---such as FLAME expression parameters, SMPL body pose, and camera configurations---while maintaining strict temporal coherence and multi-view consistency.

Formally, at time step $t$, let  
\begin{itemize}
    \item $\phi_t$ denote the FLAME parameters capturing facial expression, jaw pose, gaze, and local head pose,
    \item $\psi_t$ denote the SMPL parameters representing global body pose, articulation, and shape,
    \item $\kappa_t$ denote camera intrinsics and extrinsics, and
    \item $I^*$ denote a single reference image encoding the target identity.
\end{itemize}
The renderer estimates a mapping
\begin{equation}
    V_t = \mathcal{R}(\phi_t, \psi_t, \kappa_t, I^*)
\end{equation}
that produces the RGB frame $V_t$ such that it satisfies the following properties:

\textbf{Identity Preservation: }
The generated frames must remain consistent with the reference identity across all poses, motions, and viewpoints, avoiding drift throughout the sequence.

\textbf{Parameter-Faithful Motion Rendering: }
Facial expressions, head movements, and body articulation must accurately reflect the supplied FLAME/SMPL parameters, enabling fine-grained control without over-smoothing.

\textbf{Multi-View Geometric Consistency:} Since the Thinker or Body Animator may dynamically specify different viewpoints, the renderer must produce output that respects camera geometry and maintains cross-view consistency.

\textbf{Temporal Stability:} The renderer must avoid flickering or stochastic drift across frames, ensuring smooth temporal evolution despite the inherent stochasticity of diffusion-based generation.

\subsubsection{Challenges}
Achieving parameter-based, identity-consistent rendering introduces several key challenges:

\begin{enumerate}
    \item \textbf{Disentangling identity from motion parameters.}
    Motion parameters encode pose and expression but contain no identity cues, requiring the renderer to integrate them stably with the identity embedding.

    \item \textbf{Bridging low-dimensional parameter spaces with image-space features.}
    FLAME and SMPL parameter spaces are compact, whereas video diffusion models operate on dense spatial embeddings, necessitating carefully designed adapters.

    \item \textbf{Maintaining cross-view coherence.}
    Diffusion models are prone to view inconsistency unless explicitly trained with multi-view objectives and camera-aware modulation.

    \item \textbf{Avoiding distribution shift across heterogeneous datasets.}
    Datasets with high-quality FLAME labels seldom include multi-view imagery, and vice versa, making it difficult to jointly learn facial control and geometric consistency.
\end{enumerate}

\subsection{Approach: AvatarDiT}

\label{sec:render}
\begin{figure}[t]
    \centering
    \includegraphics[width=0.7\linewidth]{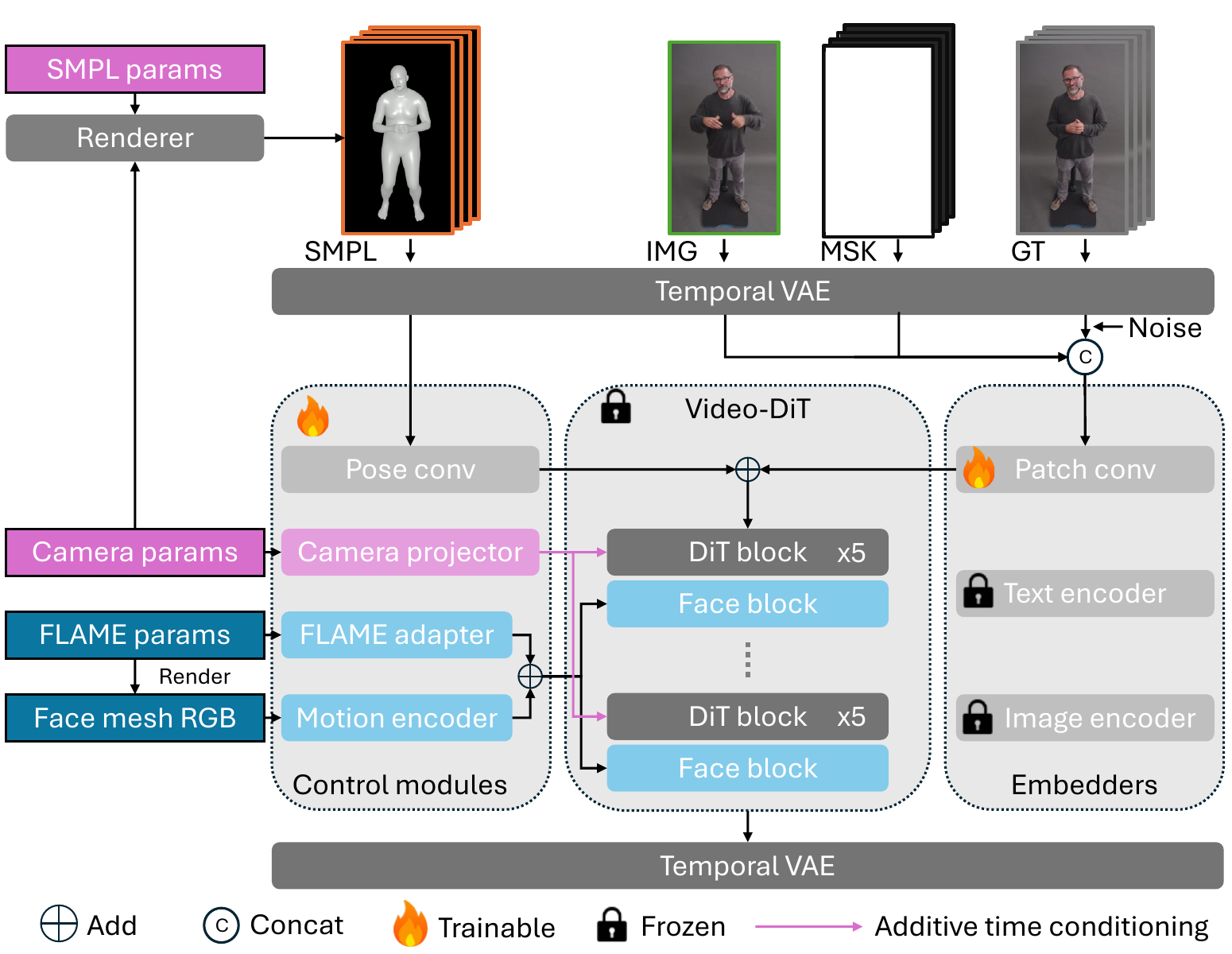}
    \includegraphics[width=1\linewidth]{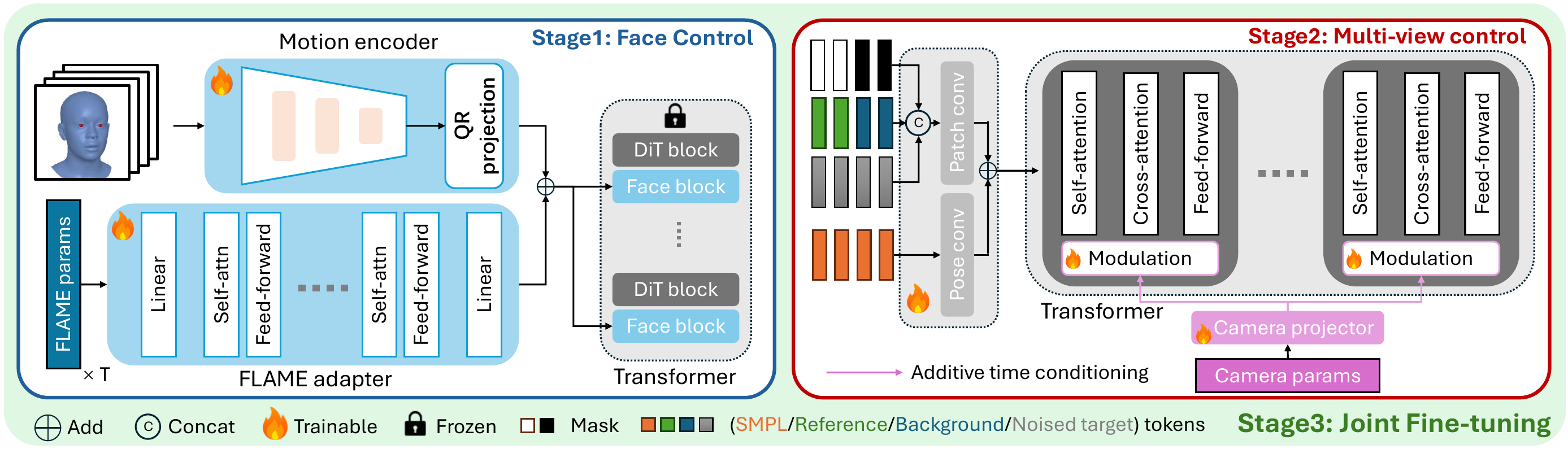}
    \caption{Overview of the proposed framework. The training process consists of 3 stages: (1) Face Control Stage, which learns parameter-driven facial motion control with FLAME-rendered mesh supervision; (2) Multi-View Stage, which enforces cross-view consistency across multiple camera parameters; and (3) Joint Stage, which integrates both to achieve coherent spatio-temporal alignment.}
    \label{fig:networkoverview}    
\end{figure}

We aim to achieve parameter-based rendering through a video Diffusion Transformer (vDiT). By leveraging the strong generative capability of the WAN model~\cite{wan-animate}, our framework supports identity-consistent multi-view generation and parameter-driven facial control. Unlike existing approaches that rely on reference images or driving videos, our method controls facial motion directly via FLAME parameters, enabling precise and disentangled manipulation. As illustrated in Figure~\ref{fig:networkoverview}, the proposed framework consists of three major components: (1) a Diffusion Transformer serving as the denoising backbone, (2) text and image encoders together with a patch convolution embedder for multi-modal conditioning, and (3) a set of control modules jointly optimized during training to inject parameter-based motion and identity information into the generation.

Given the distinct nature of facial motion control and multi-view generation, and the difficulty of collecting data that jointly captures both modalities, we adopt a three-stage training pipeline as shown the lower part in Figure~\ref{fig:networkoverview}. 

\begin{itemize}
    \item \textbf{Face control stage:} The FLAME adapter and motion encoder are trained to enable parameter-based facial control by replacing the original RGB conditioning with FLAME parameters. 
    \item \textbf{Multi-view control stage:} To address view inconsistency and identity drift, we introduce a cross-view training strategy that enforces geometric and appearance coherence across camera poses. 
    \item \textbf{Joint fine-tuning stage:} All modules are jointly optimized to close the distribution gap and achieve unified, identity-consistent generation.
\end{itemize}

For each training stage, we first train at a resolution of $512\times768$, and then scale up to $720\times1280$ to align with the resolution setting of WanAnimate. As illustrated in Figure~\ref{fig:teaser}, the control signals in our framework can be derived from multiple modalities, such as video, audio, or text, enabling flexible and multimodal conditioning. During inference, these aggregated parameter signals are injected into the AvatarDiT alongside a reference image to synthesize identity-consistent, parameter-driven human videos. The detailed configurations and objectives of each training stage are described in the following subsections.

\subsubsection{Parameter-based face control}
Wan Animate~\cite{wan-animate} effectively controls facial motion using RGB image inputs. However, this image-based design constrains its flexibility and generalization. To overcome this limitation, instead of extracting motion embeddings from face images~\cite{wan-animate,luo2025dreamactor}, we employ FLAME parameters \cite{FLAME:SiggraphAsia2017,EMOCA2021,DECA2021} to control facial movement. This parameter-based representation allows more general and adjustable control over facial expressions and motion, as FLAME parameters can explicitly disentangle motions from other visual attributions and are widely used as middle representations for motion extraction and facial motion generation~\cite{MICA2022,chu2024gagavatar,paraperas2025arc2face_exp}.

We adopt a 4-layer Transformer adapter to map the 112-D FLAME parameters into a 512-D \emph{face-motion embedding} space. In FLAME, the parameter vector is defined as $\phi = [\mathbf{e};\, \mathbf{r}_{\text{jaw}};\, \mathbf{r}_{\text{gpose}};\, \mathbf{r}_{\text{leye}};\, \mathbf{r}_{\text{reye}}]$, where $\mathbf{e}\in\mathbb{R}^{100}$ denotes expression coefficients and $\mathbf{r}_{\text{jaw}}, \mathbf{r}_{\text{gpose}}, \mathbf{r}_{\text{leye}}, \mathbf{r}_{\text{reye}}\in\mathbb{R}^{3}$ are axis-angle local poses (12-D total), for a total of 112 dimensions. The adapter produces $A(\phi)\in\mathbb{R}^{512}$ and injects it via element-wise residual addition into the image-derived motion embedding $E_{\text{face}}(I)$ extracted by a WanAnimate-pretrained face-motion encoder:
\begin{equation}
    z \;=\; E_{\text{face}}(I) \;+\; A(\phi).
\end{equation}
Here, $E_{\text{face}}(I)$ acts as an implicit constraint on $A(\phi)$. Without such supervision, cross-modal/scale mismatches, partial non-overlap of factors, and temporal ambiguity make semantic alignment extremely challenging.

\begin{figure}[t]
    \centering
    \includegraphics[width=0.7\linewidth]{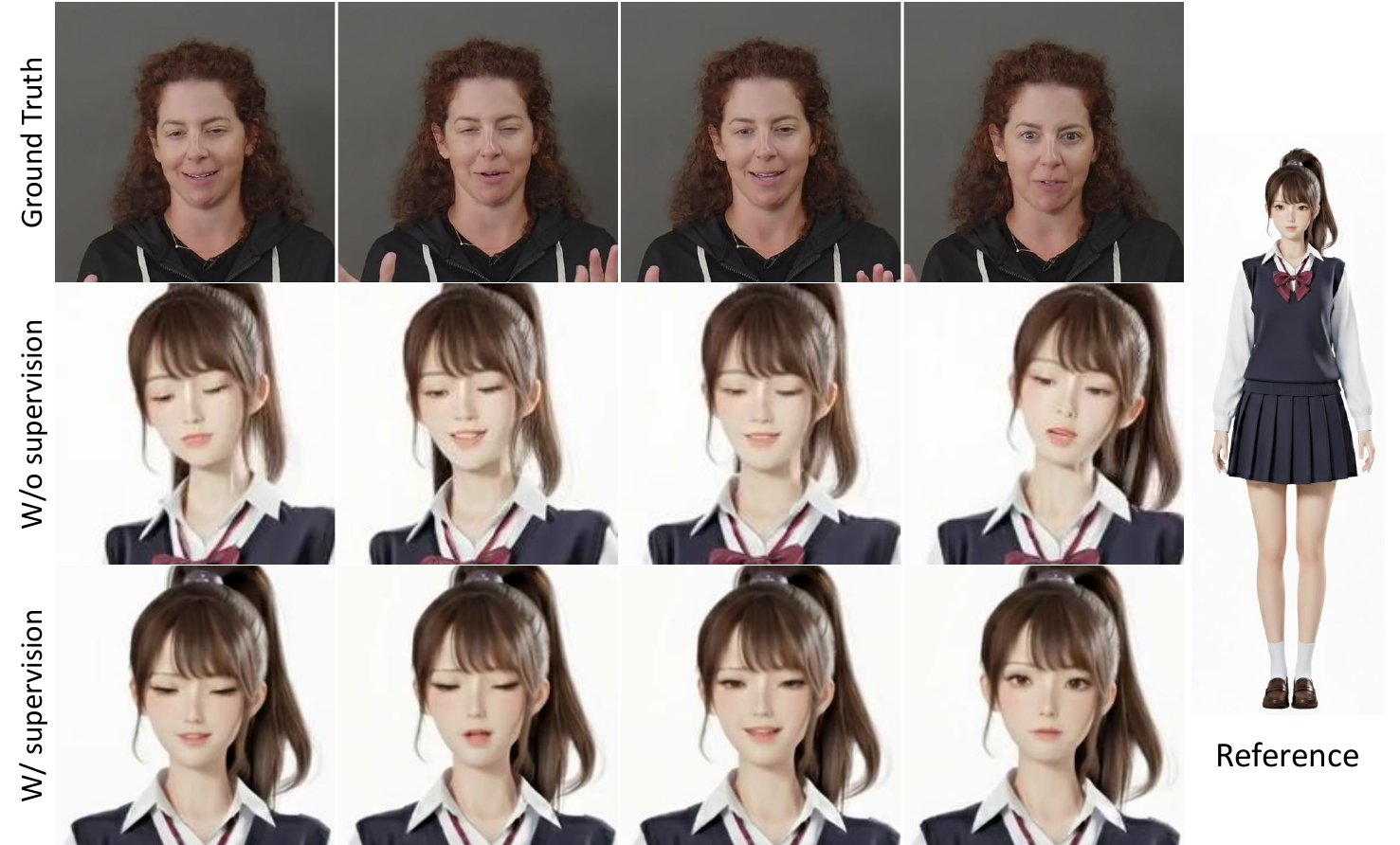}
    \caption{Ablation study of the supervision embeddings used to train the FLAME adapter. Shown are crops from our full-body generation results using on-the-fly reference conditioning.}
    \label{fig:facial-training}
\end{figure}
To enable fully parameter-based control, we jointly optimize the FLAME adapter and the motion encoder.
Since the motion encoder is pre-trained on human-face RGB images, we adapt it to also accept FLAME-rendered mesh RGB. After optimization, the framework can drive facial motion using the face mesh, the FLAME parameters, or both.

This training strategy is essential for embedding FLAME parameters into the same latent space as the original facial motion representations. An ablation study will be given in Section 4 to demonstrate its necessity. This design not only enables consistent embedding alignment but also results in comparable face control compared to the RGB-based approach adopted by Wan Animate. Similar to IP-Adapter \cite{ye2023ip-adapter}, though this training is performed on a specific dataset, the adapter can be generalized to various inputs out of the distribution.

\subsubsection{SMPL-driven Multi-view control}
A common approach in existing human animation systems for introducing motion control is to use OpenPose~\cite{openpose}, a keypoint-based modality, as control signals~\cite{wan-animate,zhang2025mimicmotion,vace}. However, OpenPose outputs are sparse and highly abstract, making it difficult to infer accurate 3D information from its RGB image. 

To enable more precise control over both synthesized motion and camera view, we instead employ SMPL~\cite{smpl-tog}-based RGB renderings as control signals. These renderings are generated from SMPL parameters together with camera poses, allowing our framework to be driven entirely by 3D controllable parameters without relying on input videos.

We basically follows the input formulation we utilize the input formulation of Wan-Animate, whose denoising target comprises chunks of \emph{reference latents}, \emph{temporal latents}, and an \emph{environmental latent}. This formulation naturally helps maintain cross-view consistency when reference latents are extracted from multi-view images. In our multi-view training, we randomly select $1$--$5$ reference frames from different views and encode them into the latents as original WanAnimate. Yet, we introduce a trainable module to improve multi-view consistency.

Additionally, we finetune the modulation layers at each DiT block to introduce a \emph{camera-based shifting}, analogous to the timestep-embedding shifting in Wan-Animate. Let the camera parameters (intrinsics/extrinsics) be embedded into three channel-wise scalar vectors $e_0,e_1,e_2 \in \mathbb{R}^{C}$ via linear modulations. Let $\mathrm{FFN}(\cdot)$ denote the DiT feed-forward MLP, and let $\mathrm{Norm}(\cdot)$ denote the channel-wise normalization in the block. Denote by $z^{(\mathrm{ca})}$ the cross-attention output within the same DiT block. We adopt:
\begin{equation}
\label{eq:cam-shift}
z_{\mathrm{out}} = z^{(\mathrm{ca})}
+ \mathrm{FFN}\!\Big(\mathrm{Norm}(z^{(\mathrm{ca})}) \odot (1 + e_1) + e_0\Big)\odot e_2,
\end{equation}
where $\odot$ denotes channel-wise multiplication.

\textbf{Joint training.} After independently training the FLAME adapter and multi-view adapter on respective datasets, we need to align their distributions to avoid out-of-distribution (OOD) artifacts and achieve a better performance. We perform a   existing open-source datasets cannot satisfy the requirement. Therefore, we organize a small fine-tuning dataset to train both adapters jointly.

\subsubsection{Dataset for Joint Fine-tuning}
To reconcile the distribution gap between the facial-control and multi-view stages, and to ensure that the final renderer can simultaneously handle parameter-driven facial motion and cross-view geometric consistency, we construct a three-stage training curriculum, each paired with a dataset whose supervision signals match the requirements of that stage.
In the face-control stage, we train the FLAME adapter and motion encoder on the \emph{Seamless Interaction} dataset~\cite{seamless-interaction}. We labeled 40  Per-frame FLAME parameters (pose, expression, jaw, and weak-perspective camera) are obtained with an off-the-shelf FLAME fitter, yielding paired RGB–FLAME supervision. For multi-view training, we use \emph{MVHumanNet}~\cite{Xiong2024MVHumanNet}, a calibrated multi-camera human dataset; synchronized frames from distinct views are sampled to form positive cross-view tuples, driving view-consistent feature learning. In the joint training stage, we curate an identity-disjoint set of short sequences that simultaneously provide synchronized multi-view imagery and frame-aligned FLAME labels, allowing the model to reconcile facial control with view geometry. Across stages, we follow official train/val splits where available (otherwise performing identity-level splits), standardize frame rate and preprocessing (face-centric cropping and mild color/temporal jitter), and train at $512\times768$ before upscaling to $720\times1280$ to match our setting.

\begin{figure}[tbhp]
    \centering
    \includegraphics[width=0.7\linewidth]{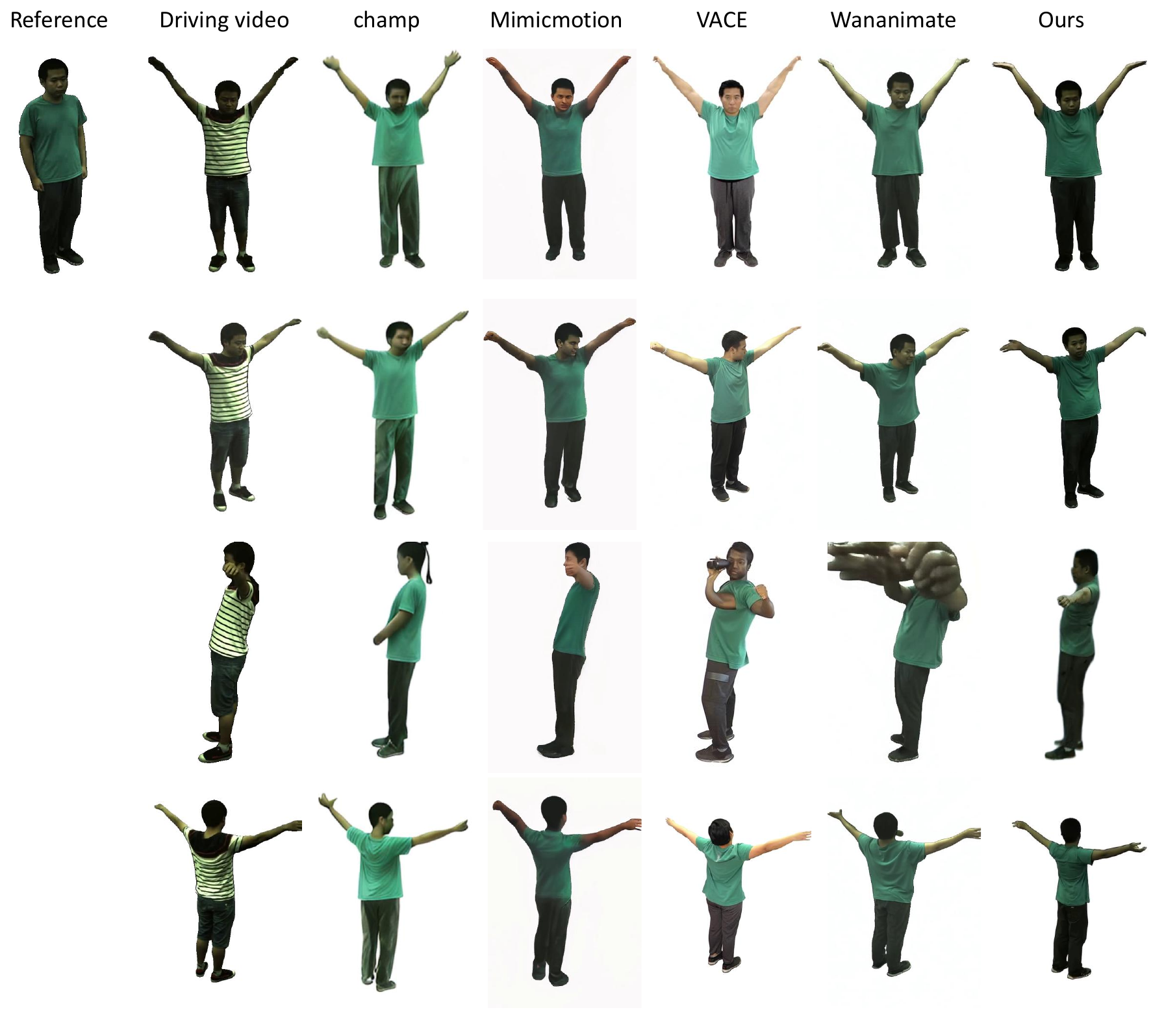}
    \caption{Qualitative comparison for multi-view results with different camera parameters.}
    \label{fig:multiview}
\end{figure}

\subsection{Results}
The DiT-based Rendering module converts parameterized facial and body motion into identity-consistent, high-fidelity human videos under dynamic pose and camera conditions. By relying exclusively on FLAME and SMPL parameters with explicit camera control, the renderer achieves stable, controllable synthesis without requiring reference driving videos.

As shown in Figure 14, our method demonstrates superior multi-view consistency compared to existing approaches, maintaining coherent facial identity, body shape, and appearance across diverse viewpoints. In contrast, prior methods frequently exhibit identity drift or view-dependent artifacts when camera parameters change. The proposed parameter-based facial control further enables accurate rendering of fine-grained expressions while preserving temporal stability.

Overall, the DiT-based Rendering module provides a robust visual realization layer for interactive digital humans, ensuring faithful motion rendering and strong identity preservation across long sequences. Additional quantitative and user-study evaluations are provided in Section 7.5.

\section{Thinker}
\begin{figure}
    \centering
\includegraphics[width=\linewidth]{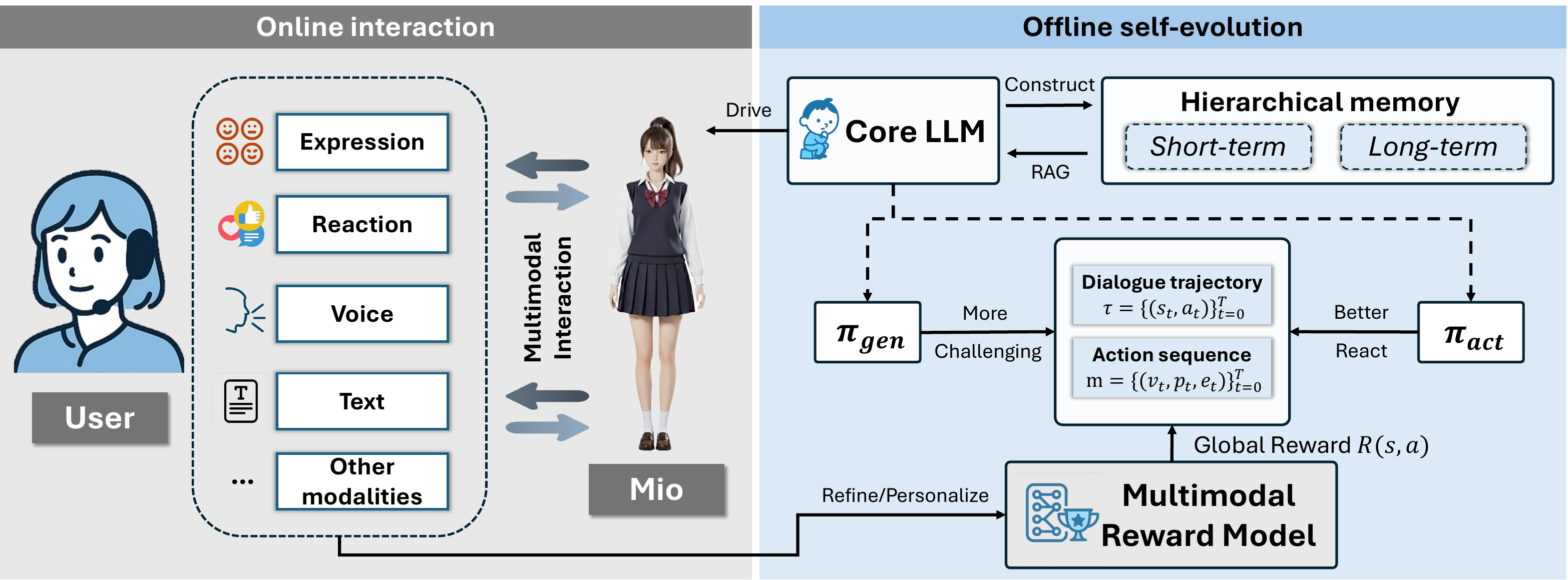}
    \caption{\textbf{The overview of the Thinker module in Mio.} (Left) Online Interaction: The Core LLM functions as the central orchestrator, driving Mio's multimodal expressions (voice, reaction, text) in real-time. It leverages a Hierarchical Memory system—comprising a short-term context buffer and a long-term Diegetic Knowledge Graph—accessed via Story-Time-Aware RAG to ensure narrative consistency and prevent spoiler leakage.
(Right) Offline Self-Evolution: To refine persona fidelity without manual annotation, the system employs a competitive self-training loop. A Generative Policy ($\pi_{gen}$) constructs challenging interactive scenarios (dialogue trajectories) to probe the agent's weaknesses. The Actor Policy ($\pi_{act}$) then optimizes its responses based on feedback from a Multimodal Reward Model, which decomposes global user satisfaction signals into fine-grained action rewards $R(s,a)$.}
    \label{fig:thinkeroverview}
\end{figure}
\subsection{Task Description}
We formulate the high-level cognitive processing of the digital human as a continuous, multimodal context-response generation problem. The Thinker functions as the central orchestrator, tasked with mapping a stream of multimodal sensory inputs to a coherent set of unified instructions for the embodiment modules (Talker, Facial Animator, and Body Animator). 

Let $I_{t}$ denote the multimodal input at time $t$, comprising the user's textual utterances, acoustic features, and visual cues. Let $M$ represent the agent's internal state, comprising both short-term context and long-term narrative knowledge. The Thinker estimates the policy:$$A_{t} = \pi(I_{t}, M)$$where $A_{t}$ is the unified action plan consisting of dialogue content, emotional state parameters, and physical gestures. The objective is to generate $A_{t}$ such that it maximizes personality coherence, narrative causality, and user engagement, driving emergent behaviors that go beyond pre-scripted rules.

To achieve genuine interactive intelligence within a specific narrative context, we must overcome several fundamental challenges. First, narrative causality and spoiler leakage. Standard retrieval-augmented generation (RAG) systems utilize temporally flat vector databases~\cite{lewis2020retrieval,huang2024egoexolearn}. In a narrative-driven setting, this leads to incoherence or spoilers, where the agent accesses information from future events (e.g., $t_{retrieved} > t_{current}$), violating the causal logic of the story~\cite{kim2025chronological,Liu2024how}. The system must maintain a strict ``Narrative-Present'' boundary while retaining access to deep historical lore. Second, temporal credit assignment in dialogue. Reinforcement learning for dialogue suffers from sparse reward signals. A user's satisfaction is typically expressed as a global signal $R_{Global}$ (\textit{e.g.}, a post-session rating) rather than immediate feedback for specific turns. Attributing this global signal to fine-grained local actions $a_{t}$ without manual turn-by-turn annotation is a critical hurdle for autonomous improvement. Third, data-free persona alignment. Aligning a general-purpose Large Language Model (LLM) to a specific, nuanced character persona usually requires extensive manual annotation or supervised fine-tuning~\cite{tseng2024two,peng2024quantifying}. The challenge lies in enabling the agent to autonomously refine its behavior and voice through self-evolution, rejecting out-of-character hallucinations without relying on external labeled datasets.

\subsection{Approach: NPC-Centric Embodied LLM}
At the heart of the Thinker is a specialized Large Language Model (LLM) tailored specifically for NPC embodiment and real-time interaction. We ground this NPC-centric model in the specific lore, context, and personality constraints of its designated character, using the following techniques.

To provide Mio with a coherent sense of self and a consistent memory, we have implemented a hierarchical memory architecture~\cite{hatalis2023memory,park2023generativeagents}. This system is the core mechanism for grounding the character's identity, ensuring that its behavior and dialogue remain consistent across long-term interactions while strictly adhering to narrative causality.

The architecture is composed of two distinct but interconnected tiers:

\textbf{1. Short-term Memory (Context Buffer):} This tier functions as a high-speed, volatile conversational buffer. It stores the immediate context of an interaction, including recent utterances, dialogue history, and currently active goals, allowing Mio to track the moment-to-moment flow of conversation.

\textbf{2. Long-term Memory (Diegetic Knowledge Graph):} To address the issues of narrative incoherence and spoiler leakage common in standard vector databases, we implement the long-term memory as a Diegetic Knowledge Graph~\cite{guo2024lightrag,huang2020improving}. Rather than a flat semantic index, this graph structures foundational memories, personality traits, and world lore into entities (nodes) and relations (edges). Critically, every element in this graph is explicitly tagged with a story-time coordinate ($t$), anchoring facts to the specific moment they occur in the narrative timeline.

At inference time, these two tiers work in synergy via a Story-Time-Aware Retrieval mechanism. The system executes a dual-level retrieval pipeline: it first performs semantic search on graph nodes to capture specific entities (low-level details) and then on edges to capture thematic relationships (high-level context). Crucially, this retrieval is governed by a Narrative-Present gate. If Mio is currently situated at time $t_{current}$, the gate rigorously filters out any memory nodes where $t_{node} > t_{current}$. This architectural constraint guarantees that Mio cannot access or leak information about events she should not have known, ensuring deep narrative consistency and preventing frame-breaking errors regardless of user prompting.

As for model training, self-evolution is the learning process that allows Mio to adapt and improve its interactive capabilities over time. This is accomplished through an autonomous learning loop that uses feedback from real user interactions to refine the LLM. This process incorporates Deep Persona Alignment (DPA) techniques to ensure high fidelity to the character's voice and values without relying on manual annotation.

\subsubsection{Multimodal reward model}
\label{sec:self_evolution}
The Multimodal Reward Model is the core component that provides the ground-truth feedback signal for learning. It addresses the fundamental temporal credit assignment problem in dialogue, which is the challenge of attributing a single, sparse global reward (e.g., a user's post-session satisfaction rating) to the specific, fine-grained local actions that caused it~\cite{ouyang2022training}. To solve this without requiring manual turn-by-turn annotation, we employ a framework where a Large Language Model (LLM) decomposes the global signal by interpreting the user's implicit multimodal cues.

For a given dialogue trajectory, $\tau = \{(s_t, a_t)\}_{t=0}^{T-1}$, we first extract a time-aligned multimodal feature vector, $m_t$, representing the user's reaction within each state, $s_t$. This vector concatenates features from different modalities:
\begin{equation}
m_t = [v_t , p_t , e_t]
\end{equation}
where $v_t$ is the visual feature vector (gaze, smile, motion probabilities), $p_t$ is the prosodic vector (pitch, intensity, jitter), and $e_t$ is the textual semantic embedding.

These numerical vectors are then transformed by a function, $f_{\text{desc}}$, into natural language descriptors, $d_t = f_{\text{desc}}(m_t)$. This step creates a text-based representation of non-verbal cues (e.g., ``User's pitch was high and they were smiling'') that is compatible with an LLM.

We then employ a powerful, frozen LLM as a zero-shot reward decomposition oracle, which we denote as the function $M_{\text{oracle}}$. This oracle takes the augmented trajectory, $\tau' = \{(s_t, a_t, d_t)\}_{t=0}^{T-1}$, and the final scalar global reward, $R_{\text{Global}}(\tau)$, as input. Its function is to output a sequence of turn-level rewards, $\{R_t\}$, for each of Mio's actions, $a_t$:
\begin{equation}
\{R_t\}_{t=0}^{T-1} = M_{\text{oracle}}(\tau', R_{\text{Global}}(\tau))
\end{equation}
The prompt given to $M_{\text{oracle}}$ includes a soft constraint that the sum of the decomposed local rewards should be faithful to the global signal. This objective can be expressed as:
\begin{equation}
\sum_{t=0}^{T-1} R_t \approx R_{\text{Global}}(\tau)
\end{equation}
The resulting local rewards, $R_t$, which are now grounded in the user's implicit multimodal reactions, are stored as the ground-truth feedback signal.

\subsubsection{Data-free self training}
To enable our Mio to autonomously refine its capabilities, we employ a data-free training protocol where the model improves through competitive self-play~\cite{lee2023rlaif,rafailov2023direct,yuan2023rrhf,huang2023weakly}. For this reinforcement learning loop to function, the agent must have a defined action space through which it can interact and a computable reward function to evaluate the quality of those actions. This structure provides the necessary feedback for iterative improvement without requiring external datasets.

We instantiate this as a game wherein the model operates under two opposing policies:

\begin{enumerate}
    \item \textbf{A Scenario-Generative Policy ($\pi_{\text{gen}}$):} In this mode, the model is tasked with generating complex and challenging interactive scenarios. The objective of this policy is to create situations designed to probe for weaknesses in the agent's reasoning, emotional appropriateness, or personality consistency. To ensure robustness, this policy employs three specific generation strategies:
    \begin{itemize}
        \item \textit{Timeline-Cycled Questioning:} Extracting events from the Diegetic Knowledge Graph to probe knowledge across the entire narrative arc.
        \item \textit{Randomized Tone Generation:} Simulating various user emotional states (e.g., tense, sarcastic, angry) to test emotional appropriateness.
        \item \textit{Varied Intent Simulation:} Generating prompts with distinct goals, such as information seeking, challenging, or negotiating.
    \end{itemize}
    \item \textbf{An Interactive Actor Policy ($\pi_{\text{act}}$):} In this mode, the model embodies Mio. Its action space consists of generating a holistic, multi-part plan that includes dialogue, an emotional state, and physical gestures. This policy is trained to respond optimally to the generated scenarios.
\end{enumerate}

To provide a high-quality training signal without human labels, we generate \textbf{Synthetic Preference Pairs} for each scenario. A teacher model creates a positive sample ($o_{pos}$), representing an ideal, in-character response, and a negative sample ($o_{neg}$), representing a flawed response exhibiting persona drift or frame-breaking behavior.

The training process follows a minimax objective optimized via Group Relative Policy Optimization (GRPO)~\cite{guo2025deepseek}. The actor policy aims to maximize a composite reward function that specifically targets persona fidelity:
\begin{equation}
\min_{\pi_{\text{gen}}} \max_{\pi_{\text{act}}} 
\mathbb{E}_{s,a} \bigl[ \mathcal{R}(s, a) - \beta D_{KL}\!\bigl(\pi_{\text{act}}(\cdot|s) \,\|\, \pi_0(\cdot|s)\bigr) \bigr]
\end{equation}
Here, the total reward $\mathcal{R}(s, a)$ incorporates the multimodal feedback $R_t$ defined previously, alongside a specialized \textbf{Persona-Based Reward} $r(a)$ derived from the synthetic pairs:
\begin{equation}
r(a) = w_{sim}\bigl(\text{sim}(a, o_{pos}) - \text{sim}(a, o_{neg})\bigr) + w_{form}\bigl(\text{form}(a, o_{pos}) - \text{form}(a, o_{neg})\bigr)
\end{equation}
In this equation, $\text{sim}(\cdot)$ represents semantic similarity between embedding representations, and $\text{form}(\cdot)$ represents surface-form similarity based on edit distance. By maximizing the margin between the generated action $a$ and the positive/negative anchors (weighted by $w_{sim}$ and $w_{form}$), Mio learns to internalize the nuanced voice and values of the character while actively rejecting out-of-character behaviors.

\section{Benchmark}
\subsection{Evaluation Metrics}

\paragraph{Talker.} The talker evaluation consists of two parts: (1) the speech reconstruction evaluation of the Kodama-Tokenizer and (2) the zero-shot TTS evaluation of the Kodama-TTS model. We measure the model's capability of reconstructing and generating intelligible, natural, and speaker-consistent speech in comparison with baseline models.  The evaluation encompasses both objective metrics including DNSMOS (audio cleanliness)~\cite{reddy2021dnsmos}, SIM (Speaker Similarity), calculated via a WavLM-Large speaker embedding model that measures the fidelity of the reconstructed speaker identity, STOI (Short-Time Objective Intelligibility) that assesses speech clarity, particularly in noisy conditions, PESQ (Perceptual Evaluation of Speech Quality) that evaluates the amount of distortion, in both narrowband (NB) and wideband (WB) settings, and WER/CER (Word/Character Error Rate) that measures the pronunciation accuracy. Subjective metrics reflecting human preference is assessed via a user study, which is represented by Mean Opinion Score (MOS) for Naturalness and Speaker Similarity.

\paragraph{Facial Animator.}
Our metrics evaluate both speaking accuracy and listening naturalness.
\noindent \textit{a) Speaking.} To assess lip synchronization quality, we use Lip Vertex Error (LVE) \citep{richard2021meshtalk}, which measures the maximum per-frame lip-vertex error. To assess overall facial accuracy, we use Mean Head Distance (MHD), the average distance across all head vertices. To assess upper-face motion consistency, we use Upper-face Dynamic Deviation (FDD) \citep{codetalker2023}. To assess temporal pose dynamics, we use Pose Dynamic Deviation for head pose (PDD) and jaw pose (JDD).
\noindent \textit{b) Listening.} For listening behaviors, we compute FDD, PDD, and JDD on listening segments to assess whether the generated motion dynamics follow the ground-truth distribution. We additionally report FID on FLAME expression and pose parameters to assess distributional naturalness.

\paragraph{Body Animator.}
To validate the Body Animator, we establish a comprehensive benchmark focusing on both motion quality and streaming responsiveness. We utilize the \textbf{HumanML3D} dataset~\cite{guo2022generating} as the standard testbed, where our module achieves an FID of \textbf{0.057} and R-Precision@3 of \textbf{0.810}, matching state-of-the-art offline models. Crucially, to evaluate performance under real-time constraints, we extend the benchmark to a streaming setting using the \textbf{BABEL} dataset~\cite{punnakkal2021babel}. In this streaming protocol, FloodDiffusion records a Peak Jerk (PJ) of \textbf{0.713} and Area Under Jerk (AUJ) of \textbf{14.05}, significantly outperforming existing streaming baselines (e.g., MotionStreamer with PJ 0.912) in terms of \emph{Transition Smoothness}, ensuring stable low-latency motion for the Thinker-Renderer pipeline.

\paragraph{Thinker.} To clearly measure our system’s Interactive Intelligence, we evaluate the Thinker module inside a detailed literary simulation built from Jules Verne’s \textit{Twenty Thousand Leagues Under the Sea}. This setting offers a fixed reference for character personality, knowledge, and story logic, all of which open-ended environments cannot guarantee. For example, suppose the architecture can maintain the distinctive voice of a character (such as Captain Nemo) and still follow the novel’s strict narrative causality. In this case, it demonstrates that the Thinker can sustain coherent personalities and reason effectively within context. Our evaluation uses several complementary metrics: the CharacterBox protocol~\cite{wang2025characterbox} to assess behavioral fidelity, custom adversarial tests to evaluate robustness against frame-breaking, and user studies to measure sustained engagement over time.

\paragraph{Renderer.} To validate our AvatarDiT Renderer, we focus on three complementary aspects: identity preservation, multi-view geometric consistency, and perceptual video quality. Identity consistency is measured using face-embedding cosine similarity and CLIP image similarity, capturing both biometric fidelity and high-level appearance alignment. Multi-view geometric consistency is assessed using LPIPS between views rendered from identical motion states. Perceptual quality is evaluated with SSIM for structural fidelity and FID/FVD for distribution-level realism of images and videos. Temporal stability is quantified via frame-to-frame perceptual variation to penalize flickering. In addition to quantitative metrics, we also conduct a user preference study comparing realism, identity consistency, and overall visual quality.

\subsection{Talker Evaluation}

\begin{table}[t]
\centering
\caption{Evaluation on the speech reconstruction task.}
\label{tab:codec_eval}
\begin{tabular}{lcccccc}
\hline
\textbf{Dataset / Model} & \textbf{BPS} & \textbf{Frame Rate/s} & \textbf{SIM} & \textbf{STOI} & \textbf{PESQ-NB} & \textbf{PESQ-WB} \\
\hline
\multicolumn{7}{l}{\textbf{LibriTTS test-clean}} \\
Kodama-Tokenizer          & 1k          & 12.5 & 0.81 & \textbf{0.94} & \textbf{3.26} & \textbf{2.67} \\
XY-Tokenizer    & 1k          & 12.5          & \textbf{0.83} & 0.91           & 3.00          & 2.41          \\
XCodec2.0       & 0.8k & 50          & 0.82 & 0.91           & 3.03          & 2.43          \\

\hline
\multicolumn{7}{l}{\textbf{Seed-TTS-Eval-ZH }} \\

XY-Tokenizer  & 1k   & 12.5          & \textbf{0.87} & 0.90          & 2.88          & 2.24          \\
XCodec2.0     & 0.8k & 50            & 0.81 & 0.89          & 2.69          & 2.10          \\
\textbf{Kodama-Tokenizer}          & 1k   & 12.5 & 0.84 & \textbf{0.91} & \textbf{3.07} & \textbf{2.60} \\
\hline
\multicolumn{7}{l}{\textbf{Seed-TTS-Eval-EN }} \\

XY-Tokenizer        & 1k   & 12.5          & 0.82          & 0.90          & 2.69          & 2.14          \\
XCodec2.0           & 0.8k & 50            & \textbf{0.89} & 0.89          & 2.57          & 2.01          \\
\textbf{Kodama-Tokenizer}       & 1k   & 12.5 & 0.81          & \textbf{0.91} & \textbf{2.88} & \textbf{2.35} \\
\hline
\multicolumn{7}{l}{\textbf{JSUT }} \\

XY-Tokenizer        & 1k   & 12.5          & 0.60          & 0.90          & 2.28          & 1.89          \\
XCodec2.0           & 0.8k & 50            & 0.71 & 0.90          & 2.21          & 1.84          \\
\textbf{Kodama-Tokenizer}     & 1k   & 12.5 & \textbf{0.75}          & \textbf{0.95} & \textbf{3.37} & \textbf{2.81} \\
\hline
\end{tabular}

\caption{Evaluation on zero-shot TTS task.}
\label{tab:tts_eval}
\begin{tabular}{lccccc}
\hline
\textbf{Model} & \textbf{UTMOS} & \textbf{DNSMOS} & \textbf{Error Rate} $\downarrow$ & \textbf{N-MOS} & \textbf{SS-MOS} \\
\hline
\multicolumn{6}{l}{\textbf{Seed-TTS-Eval-EN (WER)}} \\
MOSS-TTSD   & 3.58          & 3.01          & 8.61\%          & \textbf{4.00} & 3.73 \\
Higgs       & 3.71          & 3.09          & 2.41\% & 3.63          & 3.90 \\
\textbf{Kodama-TTS}  & \textbf{4.06}          & \textbf{3.14} & \textbf{1.54}\%          & 3.85          & \textbf{4.03} \\
\hline
\multicolumn{6}{l}{\textbf{Seed-TTS-Eval-ZH (CER)}} \\
MOSS-TTSD   & 2.92 & 3.20          & 2.94\%          & \textbf{3.83} & \textbf{3.93} \\
Higgs       & 2.67          & 3.22 & 2.43\% & \textbf{3.83} & 3.83 \\
\textbf{Kodama-TTS}  & \textbf{3.09}          & \textbf{3.30}          & \textbf{1.36}\%          & 3.73          & 3.78 \\
\hline
\multicolumn{6}{l}{\textbf{CV3-Eval-JA (CER)}} \\
MOSS-TTSD   & 2.21          & 3.02          & 344.5\%         & 3.20          & 3.65 \\
Higgs       & 2.69 & \textbf{3.05} & 94.2\%          & 1.55          & 2.13 \\
\textbf{Kodama-TTS}  & \textbf{2.98} & \textbf{3.15}          & \textbf{4.74\%} & \textbf{4.20} & \textbf{3.95} \\
\hline
\end{tabular}
\end{table}

\subsubsection{Speech Reconstruction Performance} 
The reconstruction performance evaluates the tokenizer's ability to preserve semantic and acoustic information within a low-bitrate constraint. Table~\ref{tab:codec_eval} presents the comparative results across multiple datasets. We use the test-clean subset of LibriTTS corpus to test the models' performance in reconstructing clean speech, and Seed-TTS-Eval~\cite{anastassiou2024seed} for a more in-the-wild setting. For Japanese, we use the JSUT~\cite{sonobe2017jsut} corpus. The Kodama-Tokenizer demonstrates a significant advantage in audio quality and intelligibility. It consistently leads in perceptual quality, achieving a PESQ-NB of 3.26 on LibriTTS and 3.07 on Seed-TTS-Eval-ZH, substantially outperforming baselines like XY-Tokenizer (3.00 and 2.88, respectively) and XCodec2.0. The improvement is most pronounced in the JSUT dataset, where Kodama-Tokenizer achieves a PESQ-NB of 3.37 compared to XY-Tokenizer's 2.28. Furthermore, the model excels in intelligibility, maintaining STOI scores above 0.91 across all test sets, with a peak of 0.95 on JSUT, ensuring robust speech comprehension.

We acknowledge a current limitation regarding speaker similarity. As observed in the LibriTTS benchmark, Kodama-Tokenizer scores approximately 0.81, which is slightly lower than XY-Tokenizer (0.83), and the gap is larger on Seed-TTS-Eval datasets in English (0.81 vs XCodec2.0's 0.89) and Chinese (0.84 vs XY-Tokenizer's 0.87), although it has surpassed the counterparts in Japanese, marking the best with a SIM score of 0.75. This indicates a trade-off where the model prioritizes reconstruction clarity and naturalness over absolute speaker embedding fidelity in noiser settings, because the Seed-TTS-Eval dataset is sampled form the Common Voice corpus~\cite{ardila2020common}, which is collected through crowdsourcing and has a varying recording quality. Overall, Kodama-Tokenizer offers a superior balance of compression efficiency and high-fidelity reconstruction, positioning it as a robust solution for real-time, low-latency applications where audio quality and intelligibility are paramount.

\subsubsection{Zero-Shot TTS Performance}

We evaluate the zero-shot text-to-speech capabilities of Kodama-TTS using the same Seed-TTS-Eval dataset for English and Chinese. For Japanese, since there is no standard benchmark dateset with multiple speakers, we randomly selected 2000 samples from Common Voice's \textit{cv-corpus-23.0-2025-09-05} corpus and organized the prompt audio and target texts in the same way as Seed-TTS-Eval. We benchmark against two state-of-the-art open-source models that adopt the same AR+Codec architecture: MOSS-TTSD, trained on over 1 million hours of data using the XY-Tokenizer~\cite{gong2025xy}, and Higgs~\cite{higgsaudio2025}, a massive model trained on over 10 million hours of speech. As detailed in Table~\ref{tab:tts_eval}, Kodama-TTS demonstrates remarkable performance, effectively matching the capabilities of the 10M-hour baseline in English while establishing a significant lead in Japanese.

\noindent\textbf{Objective Performance Analysis.} In English scenarios, Kodama-TTS achieves a DNSMOS of 3.13, surpassing both MOSS-TTSD (3.01) and the significantly larger Higgs model (3.09), indicating superior audio signal quality. In terms of pronunciation stability, its Word Error Rate (WER) of 2.50\% is highly competitive, nearly matching Higgs (2.41\%) and significantly outperforming MOSS-TTSD (8.61\%).

The model's advantage is most pronounced in the Japanese subset. Both baselines struggle significantly here, with MOSS-TTSD and Higgs exhibiting severe pronunciation failures (CERs of 317.53\% and 92.44\%, respectively). In contrast, Kodama-TTS maintains a much lower Character Error Rate (CER) of 32.82\%, proving it to be the only robust candidate for Japanese generation among the compared models. In Chinese, while Kodama-TTS lags slightly behind the baselines in CER (6.74\% vs. 2.43\% for Higgs), it maintains comparable audio quality.

\noindent\textbf{Subjective Preference.} The Naturalness Mean Opinion Score (N-MOS) and Speaker Similarity Mean Opinion Score (SS-MOS) subjective evaluation corroborate the objective metrics. For English zero-shot generation, Kodama-TTS achieves the highest SS-MOS of 4.03, while maintaining a high Naturalness score of 3.85. In Japanese, the performance gap is distinct: Kodama-TTS dominates with a N-MOS score of 4.2 and a Similarity score of 3.95, whereas the baselines fail to produce intelligible or natural speech (e.g., Higgs scores 1.55 in Naturalness). This confirms that Kodama-TTS successfully bridges the gap with massive-scale English models while offering superior multilingual generalization in Japanese.

\subsection{Facial Animator Evaluation}

\begin{table}[t]
    \caption{
        Evaluation of speaking and listening facial motions on the Seamless Interaction \cite{seamless-interaction} test split. ARTalk* indicates that we adapt the original ARTalk \cite{artalk2025} for speak-listen generation.
    }
    \label{tab:face_main_res}
    \setlength{\tabcolsep}{1.4mm}
    \centering
    \small
    \begin{tabular}{l|ccccc|ccccc}
    \toprule[1.2pt]
    & \multicolumn{5}{c|}{Speak} & \multicolumn{5}{c}{Listen} \\
    Method      & LVE$\downarrow$ & MHD$\downarrow$  & FDD$\downarrow$  & PDD$\downarrow$ & JDD$\downarrow$ & FDD$\downarrow$  & PDD$\downarrow$ & JDD$\downarrow$ & F-FID$\downarrow$  & P-FID$\downarrow$  \\
    \midrule
    DiffPoseTalk~\citep{diffposetalk2024} & 9.48 & 2.96 & 32.66 & 7.89 & 1.40 & - & - & - & - & -\\
    ARTalk~\citep{artalk2025} & 7.46 & 2.12 & 31.64 & {7.66} & 1.19 & - & - & - & - & - \\
    ARTalk*~\citep{artalk2025} & 6.79 & 2.02 & {27.41} & 8.55 & {0.81} & {30.62} & {9.52} & {1.53} & {10.779} & {0.072}  \\
    DualTalk~\citep{peng2025dualtalk} & {6.35} & {1.95} & 37.46 & 9.70 & 1.02 & 43.58 & 10.71 & 2.02 & 13.143 & 0.079 \\
    \midrule
    UniLS (ours) & \textbf{5.83} & \textbf{1.89} & \textbf{18.41} & \textbf{4.67} & \textbf{0.71} & \textbf{17.12} & \textbf{4.75} & \textbf{0.98} & \textbf{4.304} & \textbf{0.038} \\
    \bottomrule[1.2pt]
    \end{tabular}
\end{table}

\paragraph{Quantitative Results.} \Cref{tab:face_main_res} reports an evaluation of speaking and listening facial motions.
Our facial animator shows clear improvements in lip-sync accuracy (LVE, MHD) and speech-style alignment (FDD, PDD, JDD).
These results indicate that our animator not only tracks phoneme-to-motion correspondence precisely but also captures the characteristic dynamics of speech, such as upper-face involvement and coordinated head-jaw movement.
For listening, our approach shows large improvements in distributional measures (FDD, PDD, JDD, F-FID, P-FID). 
This indicates that our animator generates diverse expressions and head movements instead of collapsing into a neutral or static listening pose.
Together, these results validate our two-stage design, showing that the model successfully learns to produce both natural listening reactions and accurate speaking expressions within a unified framework.

\paragraph{User Study.}
To thoroughly assess our method, we conducted a user study examining four key aspects of conversational facial motion: lip synchronization, facial expression naturalness, listening reaction naturalness, and head pose naturalness. 
For baseline methods that do not generate listening behaviors, the reaction naturalness category is omitted.
We adopt a pairwise comparison protocol. For each trial, videos generated by our method and a baseline model are shown side by side in a randomized order. After viewing the video pairs, participants select the result they find more natural and realistic.
We then compute the percentage of users who prefer our method in each category.

As summarized in \cref{tab:face_user_study}, our method is consistently preferred over all baselines across all aspects.
Among the 25 participants in our study, the most notable improvement appears in listening reactions, where over 90\% of participants prefer our results compared to DualTalk.
This overwhelming preference highlights the strength of our two-stage design: our model produces listening motions that are significantly more expressive, responsive, and human-like than existing methods.

\begin{table}[t]
\caption{
User study results from 25 participants. Numbers (\%) indicate the proportion of users who preferred our facial animator over each baseline. We compare performance across four aspects: lip synchronization, facial expression naturalness, listening reaction naturalness, and head pose naturalness.
}
\label{tab:face_user_study}
\setlength{\tabcolsep}{3.5mm}
\centering
\small
\begin{tabular}{lcccc}
\toprule[1.2pt]
Method & Lip Synchronization  & Expression  & Reaction  & Head Pose \\
\midrule
vs. DiffPoseTalk~\citep{diffposetalk2024}    & 55.34 & 61.65 & - & 58.25 \\
vs. ARTalk~\citep{artalk2025}                & 75.36 & 75.36 & - & 71.98 \\

vs. ARTalk*~\citep{artalk2025}           & 76.92 & 77.88 & 79.80 & 74.52 \\
vs. DualTalk~\citep{peng2025dualtalk}        & 86.06 & 90.38 & 91.35 & 89.42 \\
\bottomrule[1.2pt]
\end{tabular}
\end{table}

\subsection{Body Animator Evaluation}

We evaluate our body animator on two standard benchmarks: HumanML3D (for motion quality) and BABEL (for streaming capability).

\paragraph{Baselines.}
We compare against two primary state-of-the-art streaming baselines:
\begin{itemize}
    \item \textbf{PRIMAL} \citep{zhang2025primal}: A chunk-based diffusion model that generates motion in fixed-size segments. It suffers from high "first-token" latency because it must wait for the entire chunk to be generated before outputting.
    \item \textbf{MotionStreamer} \citep{xiao2025motionstreamer}: An autoregressive (AR) model with a diffusion head. While strictly causal, it often struggles with long-term consistency due to error accumulation and lacks the ability to refine past frames within a sliding window.
\end{itemize}
Our method combines the best of both worlds: the refinement capability of diffusion (like PRIMAL) and the low latency of causal processing (like MotionStreamer).

\paragraph{Quantitative Results.}
Table \ref{tab:quantitative_eval} summarizes the comparison against state-of-the-art methods.

\begin{table*}[thb]
\centering
\scalebox{0.85}{
\begin{tabular}{l c c c c c c c c c}
&  & \multicolumn{6}{c}{\textbf{HumanML3D}} & \multicolumn{2}{c}{\textbf{BABEL}}\\
\cmidrule(lr){3-8}\cmidrule(lr){9-10}
 & stream & R@1$\uparrow$ & R@2$\uparrow$ & R@3$\uparrow$ & FID$\downarrow$ & MM-Dist$\downarrow$ & Diversity$\rightarrow$ & PJ$\rightarrow$ & AUJ$\downarrow$\\
\midrule
Real motion & ~ & 0.511 & 0.703 & 0.797 & 0.002 & 2.974 & 9.503 & 1.100 & 41.20\\
\midrule
T2M-GPT & ~ & 0.492 & 0.679 & 0.775 & 0.141 & 3.121 & 9.722 & --& - \\
MoMask & ~ & \underline{0.521} & \underline{0.713} & \underline{0.807} & \textbf{0.045} & 2.958 & 9.677 & - & - \\
\midrule
PRIMAL & \cmark & 0.497 & 0.681 & 0.780 & 0.511 & 3.120 & \textbf{9.520} & 1.304 & 19.36 \\
MotionStreamer & \cmark & 0.513 & 0.705 & 0.802 & 0.092 & \underline{2.909} & 9.722 & \underline{0.912} & \underline{16.57} \\
\textbf{FloodDiffusion} & \cmark & \textbf{0.523} & \textbf{0.717} & \textbf{0.810} & \underline{0.057} & \textbf{2.887} & 9.579 & \textbf{0.713} & \textbf{14.05}  \\
\end{tabular}
}
\caption{\textbf{Quantitative evaluation on HumanML3D and BABEL test sets.} FloodDiffusion achieves the best R@k and MM-Dist, a competitive FID (0.057) on HumanML3D, and outperforms all streaming baselines on BABEL.}
\label{tab:quantitative_eval}
\end{table*}

Our method achieves an FID of \textbf{0.057} on HumanML3D, which is significantly better than other streaming baselines like MotionStreamer (0.092) and PRIMAL (0.511), and is on par with the best offline method MoMask (0.045). This validates that our streaming constraint does not compromise generation quality.
For streaming metrics on BABEL, we achieve the lowest Peak Jerk (PJ) and Area Under Jerk (AUJ), indicating that our transitions between different text prompts are the smoothest.

\paragraph{Ablation Studies.}
We investigated the impact of our key design choices:
\begin{itemize}
    \item \textbf{w/o Bi-directional Attention}: FID degrades to 3.377. This confirms that frames in the active window must attend to each other to resolve consistency.
    \item \textbf{w/o Lower-Triangular Schedule}: Using a random schedule (standard diffusion forcing) results in an FID of 3.883. The structured "cascading" noise is crucial for the model to learn the streaming task effectively.
\end{itemize}

\paragraph{User Study.}
We conducted a Bradley-Terry user study with 100 participants. Users preferred FloodDiffusion over PRIMAL and MotionStreamer for \textbf{Transition Smoothness} (+0.152 score) and overall preference.

\subsection{DiT Renderer Evaluation}
We evaluate the performance of the proposed framework with ablation models and baseline methods from three perspectives: face motion fidelity, multi-view identity consistency, and overall perceptual quality.
\label{sec:experiment}

\subsubsection{Ablation study}
To comprehensively evaluate the effectiveness of our design choices, we conduct a series of ablation experiments focusing on three aspects: (1) the necessity of embedding supervision for stable FLAME-based facial control, (2) the controllability and disentanglement of individual FLAME parameters, and (3) the impact of camera-based modulation on multi-view consistency. Through both qualitative and quantitative analyses, we validate that each proposed component contributes to the overall controllability, stability, and realism of the generated results.\\

\noindent\textbf{Embedding supervision.} Most existing face control models are tailored for portrait or avatar generation, where the synthesized content primarily focuses on the upper body and facial appearance. In this work, we introduce a parameter-based facial control mechanism for full-body generation through a trainable adapter. However, directly training a FLAME adapter from scratch is highly unstable and prone to losing prior motion knowledge. To enable effective control, we jointly optimize the motion encoder—which extracts facial motion embeddings from RGB inputs—with the FLAME adapter. This joint optimization accelerates convergence and successfully transfers motion priors into the parameter space. To validate the necessity of embedding supervision, we conduct an ablation in which the FLAME adapter is trained independently without the motion embedding guidance. As shown in Figure~\ref{fig:facial-training}, the ablated model fails to respond to FLAME parameter variations, confirming that embedding supervision is crucial for achieving stable and controllable facial motion generation.\\

\noindent\textbf{Fine-grained Parameter Control.} FLAME parameters are parameterized to target specific facial attributes, yielding a disentangled, interpretable control space. Using FLAME as the intermediate representation, our framework enables precise, factorized manipulation of facial motion~\cite{paraperas2025arc2face_exp}. To assess controllability, we vary the global head pose $\mathbf{r}{\text{gpose}}$ and the jaw/mouth parameter $\mathbf{r}{\text{jaw}}$ independently while zeroing all remaining FLAME coefficients. 


\noindent\textbf{Face control signal.} We then continue to evaluate the facial motion generation with different control signals. As the proposed framework is finetuned from WanAnimate~\cite{wan-animate}, we choose it as a baseline to evaluate face motion generation, as well as two ablation inference settings: (1) inference without FLAME-rendered mesh RGB, and (2) inference without FLAME parameters. These ablations mainly influence mouth-related performance metrics such as lip–audio synchronization accuracy. As shown in Figure \ref{fig:facial-training}, our full setting achieves more consistent synchronization and realistic mouth articulation, demonstrating the effectiveness of incorporating FLAME-rendered mesh guidance during both training and inference. \\

\noindent\textbf{Camera modulation.} Though SMPL shows a better expression for viewing information, the network still synthesizes mismatch results in some posterior views. To address such misalignment, we utilize camera parameters as a view control signal to notify the network of the synthesized view. A qualitative comparison is given in Figure~\ref{fig:multiview}.

\begin{table}[t]
    \centering
    \caption{Qualitative evaluation regarding multi-view consistency.}
    \renewcommand{\arraystretch}{1.05}
    \setlength{\tabcolsep}{3pt}
    \small 
    
    \begin{tabularx}{\columnwidth}{l *{6}{>{\centering\arraybackslash}X}}
        \bottomrule
        Model & SSIM$\uparrow$ & PSNR$\uparrow$ & 
        LPIPS$\downarrow$ & CLIP $\uparrow$ \\
        \midrule
        WanAnimate~\cite{wan-animate} & 0.3322 & 14.07 & 0.5277 & 0.7942\\
        VACE~\cite{vace} & 0.2674 & 9.6243 & 0.6678 & 0.7510\\
        CHAMP~\cite{zhu2024champ}  & 0.7611 & 17.42 & 0.2289 & 0.8331 \\
        \midrule
        \textbf{Ours} & 0.8134 & 16.21 & 0.2231 & 0.8693  \\
        \bottomrule
    \end{tabularx}
    \label{tab:mv-consistency}
    \vspace{-4pt}
\end{table}

\begin{table}[t]
    \centering
    \caption{Quantitative comparison with state-of-the-art methods regarding perceptual quality on our validation set.}
    \renewcommand{\arraystretch}{1.05}
    \setlength{\tabcolsep}{3pt}
    \small 
    
    \begin{tabularx}{\columnwidth}{l *{5}{>{\centering\arraybackslash}X}}
        \bottomrule
        Model & FID$\downarrow$ & SSIM$\uparrow$ & 
        PSNR$\uparrow$ & LPIPS$\downarrow$ & FVD$\downarrow$ \\
        \midrule
        WanAnimate~\cite{wan-animate} & 116.0 & 0.865 & 21.40 & 0.264 & 343.08 \\
        VACE~\cite{vace} & 114.7 & 0.778 & 14.64 & 0.315 & 340.89 \\
        Champ~\cite{zhu2024champ} & 96.22 & \textbf{0.916} & \textbf{26.62} & 0.186 & 350.48  \\
        Mimicmotion~\cite{zhang2024mimicmotion} & 77.5 & 0.905 & 25.22 & 0.334 & 226.20 \\
        \midrule
        \textbf{Ours} & \textbf{68.72} & 0.914 & 26.31 & \textbf{0.135} & \textbf{176.70} \\
        \bottomrule
    \end{tabularx}
    \label{tab:overall-perceptual}
    \vspace{-4pt}
\end{table}

\subsubsection{Comparison with Baseline}
We compare AvatarDiT against state-of-the-art controllable human animation systems, including WanAnimate~\cite{cheng2025wananimate}, VACE~\cite{jiang2025vace}, CHAMP~\cite{zhu2024champ}, and MimicMotion~\cite{zhang2024mimicmotion}. Our evaluation focuses on three axes: (1) facial motion controllability, (2) multi-view identity consistency, and (3) overall perceptual quality. Quantitative results are summarized in Tables~\ref{tab:mv-consistency}, and \ref{tab:overall-perceptual}, while qualitative examples are shown in Figure~\ref{fig:multiview}.

\paragraph{Facial motion control.}  
Table~\ref{tab:overall-perceptual} reports quantitative results. \textbf{AvatarDiT} achieves the strongest lip–audio synchronization and expression fidelity due to disentangling motion embeddings from facial appearance. Against \textbf{WanAnimate}, our FLAME-driven controller lifts PSNR from 21.40 to 27.04 and SSIM from 0.8655 to 0.9202 (↑0.0547) by excluding identity/shape factors that otherwise entangle with expression. Combining \textbf{SMPL}+ \textbf{FLAME} yields balanced face/body control; however, residual shape leakage persists from SMPL-side priors in the pre-trained motion encoder, resulting in a slightly lower SSIM and PSNR. We also benchmark against portrait-only methods~\cite{skyreels2024,xportrait-2}; because their outputs are fixed-resolution portraits, we omit the resolution-sensitive Sync score for fairness.

\paragraph{Multi-view identity consistency.}  
As shown in Table~\ref{tab:mv-consistency}, AvatarDiT substantially enhances multi-view coherence compared to existing baselines. Our method achieves the highest CLIP similarity (0.8693) and the lowest LPIPS (0.2231), demonstrating superior perceptual alignment and identity stability across viewpoints. While CHAMP attains reasonable structural consistency due to its SMPL-based control, our camera-modulated DiT blocks and multi-view training strategy further strengthen cross-view preservation. Qualitative results in Figure~\ref{fig:multiview} show that baseline methods frequently exhibit identity drift or view-dependent artifacts, whereas AvatarDiT maintains consistent appearance and geometry even under large viewpoint variations. In contrast, WanAnimate underperforms in multi-view evaluation, as it relies on OpenPose-based driving signals and is primarily designed for frontal-view synthesis.

\paragraph{Perceptual quality and overall fidelity.}  
As summarized in Table~\ref{tab:overall-perceptual}, AvatarDiT delivers a substantial boost in generation quality. WanAnimate is reliable and produces high-quality videos across diverse content; however, because it relies on 2D OpenPose for motion cues, it exhibits a strong frontal-view bias, leading to results that are inferior to CHAMP and MimicMotion. CHAMP attains the best SSIM and PSNR, benefiting from comprehensive 3D guidance, but its backbone limits overall generation fidelity, and it cannot control facial motion. Our method, which is also driven by an SMPL signal, achieves comparable SSIM/PSNR, while obtaining the best FID and FVD thanks to the strong generative capacity of WanAnimate—and, importantly, adds explicit facial-motion control.

\paragraph{Overall comparison.}  
Together, these results demonstrate that AvatarDiT outperforms existing human animation systems in both controllability and cross-view fidelity. Unlike RGB-driven or 2D-pose–driven methods, our parametric face–body representation allows for precise motion specification without reference to driving videos. The combination of FLAME-based facial control, SMPL-driven multi-view supervision, and camera-conditioned DiT layers forms a unified generative framework that achieves superior parametric precision, consistent identity preservation, and high perceptual realism.

\begin{figure}[tb]
    \centering
    \includegraphics[width=0.5\linewidth]{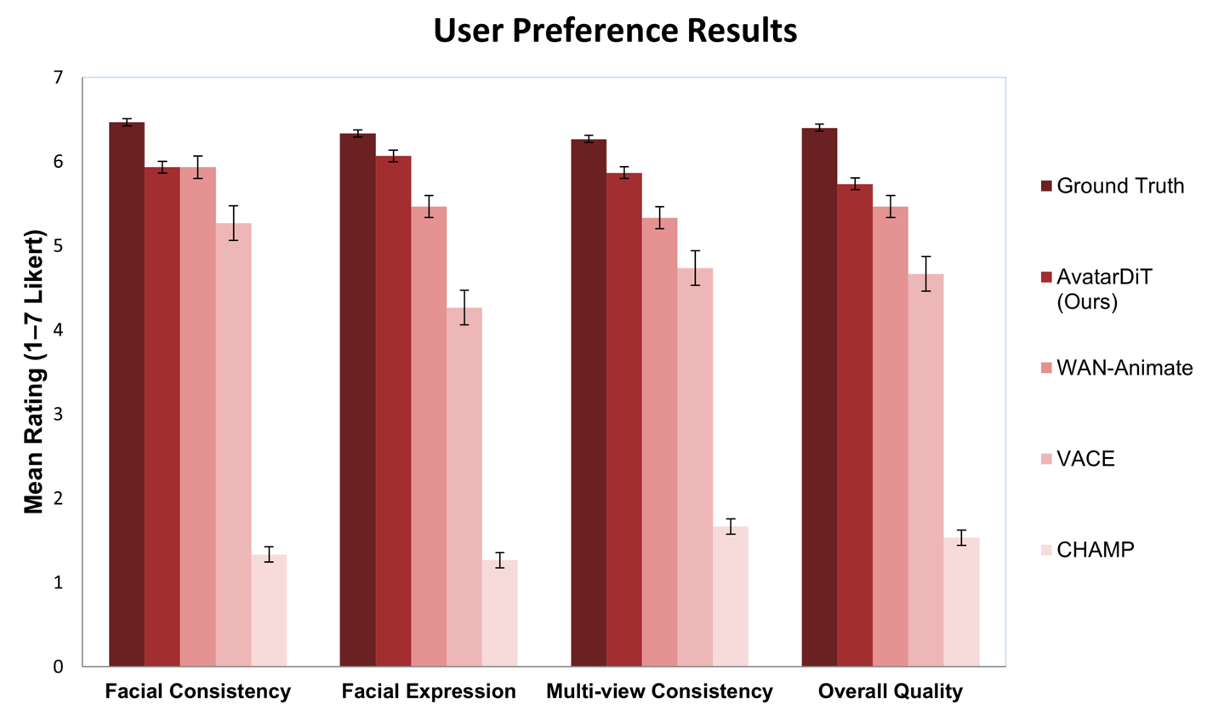}
    \caption{Results of user preference study in 7-point Likert scale.}
    \label{fig:userstudy}
\end{figure}
\subsubsection{Qualitative User Study}
To further assess the perceptual realism and controllability of AvatarDiT, we conducted a subjective user study comparing our method with three state-of-the-art baselines: WAN-Animate, CHAMP, and VACE.

Fifteen participants (aged 22–32, all with prior experience in 3D graphics or animation) were asked to evaluate 10 multi-view human-rendering videos generated by each method.
Each video was rated on a 7-point Likert scale across  four perceptual criteria: Facial Consistency, Facial Expression Accuracy, Multi-view Consistency, 
and Overall Quality.

All clips were randomized and anonymized to remove bias. Ground-truth reference renderings were included for calibration but excluded from ranking comparisons.

Figure~\ref{fig:userstudy} shows the averaged ratings across all participants. AvatarDiT achieved the highest mean scores among all generative systems in every criterion, approaching the ground-truth reference level.
Compared with WAN-Animate, our method improved Facial Expression Accuracy by +0.6 points and Multi-view Consistency by +0.5 points, demonstrating better stability and controllability. VACE achieved competitive performance but suffered from occasional side-view artifacts, whereas CHAMP was rated substantially lower due to geometric distortions and poor expressiveness under facial motion.

\subsection{Thinker Evaluation}
Our evaluation was guided by three primary research questions, which directly map to the core technical contributions of the Thinker architecture:
\begin{itemize}
    \item \textbf{RQ1: Persona Fidelity.} To what extent does our data-free self-training pipeline improve the perceived persona fidelity of a character agent compared to a standard, prompt-engineered baseline?
    \item \textbf{RQ2: System Robustness.} How does the hierarchical memory architecture affect the system's robustness against out-of-domain, frame-breaking prompts?
    \item \textbf{RQ3: Narrative Coherence.} How effectively does the diegetic knowledge graph prevent spoiler leakage compared to a standard, temporally-flat RAG system?
\end{itemize}

To provide objective and reproducible measures of system performance, we first conducted a series of automatic evaluations. We specifically evaluate four distinct configurations to isolate the impact of our contributions:
\begin{itemize}
    \item \textbf{Baseline (Prompt-Only)}: A character model driven solely by prompt engineering and equipped with a standard, temporally-flat RAG. This represents a standard off-the-shelf approach.
    \item \textbf{Self-Train Only}: Uses the models trained via our Data-Free Self-Training pipeline but relies on a standard, temporally-flat RAG over the entire novel. This condition isolates the value of the persona alignment.
    \item \textbf{Diegetic-Mem Only}: Uses the Story-Time–Aware Diegetic Memory but with character models driven only by prompt engineering. This condition isolates the value of the memory architecture.
    \item \textbf{The Full Thinker System}: The complete architecture featuring both the self-trained persona alignment and the diegetic memory constraints.
    \item \textbf{GPT-4o}: The general-purpose baseline configured with the same prompt-only persona instructions as our Baseline System.
\end{itemize}
For this experiment, we conducted our evaluation using the four main characters from \textit{Twenty Thousand Leagues Under the Sea}: Captain Nemo, Professor Aronnax, Conseil, and Ned Land.

\paragraph{Persona Fidelity via CharacterBox Benchmark}
To assess the core role-playing capabilities of our models (addressing \textbf{RQ1}), we utilized the CharacterBox benchmark~\cite{wang2024characterbox}. Evaluating role-playing is a known challenge, as simple conversational snapshots often fail to capture the nuanced behaviors and character fidelity required for authentic embodiment. CharacterBox addresses this by providing a simulation sandbox designed to generate and evaluate fine-grained character behavior trajectories within open-ended narratives, allowing for a more comprehensive assessment.

For each of the five system conditions, we evaluated all four characters. Each character evaluation was repeated 15 times to ensure statistical reliability. We evaluated all five systems across the full suite of CharacterBox metrics: Knowledge Accuracy (KA), Behavioral Accuracy (BA), Personality Traits (PT), Emotional Expression (EE), Immersion (IM), Adaptability (AD), and Behavioral Coherence (BC).

\begin{table}[t]
  \caption{\textbf{Automatic persona fidelity evaluation using the CharacterBox benchmark~\cite{wang2024characterbox}.} We compare our four system conditions (Baseline, Self-Train Only, Diegetic-Mem Only, Full Thinker System) against the GPT-4o baseline. Scores are reported (Mean $\pm$ SD) across all seven CharacterBox metrics. Scores range from 1 to 5 and higher scores are better.}
  \label{tab:characterbox}
\setlength{\tabcolsep}{5.0pt}
\centering
\resizebox{\linewidth}{!}
{
 \begin{tabular}{l|c|c|c|c|c|c|c|c} 
 \shline
  \rowcolor{mygray}
      Model & \textbf{KA} & \textbf{BA} & \textbf{EE} & \textbf{PT} & \textbf{IM} & \textbf{AD} & \textbf{BC} & \textbf{Average} \\ 
\hline
GPT-4o & 3.967$_{\pm0.97}$ & 3.683$_{\pm1.10}$ & 3.533$_{\pm0.91}$ & 3.150$_{\pm0.94}$ & 3.350$_{\pm0.97}$ & 3.500$_{\pm0.89}$ & 3.133$_{\pm0.91}$ & 3.474$_{\pm0.99}$ \\ 
Baseline & 3.417$_{\pm0.93}$ & 2.900$_{\pm0.86}$ & 3.033$_{\pm0.94}$ & 3.300$_{\pm0.91}$ & 3.317$_{\pm1.00}$ & 2.867$_{\pm0.93}$ & 3.083$_{\pm0.93}$ & 3.131$_{\pm0.94}$ \\
Self-Train Only & 3.667$_{\pm1.02}$ & 3.583$_{\pm0.91}$ & 3.550$_{\pm1.02}$ & 3.500$_{\pm0.95}$ & 3.650$_{\pm0.99}$ & 3.433$_{\pm0.98}$ & 3.700$_{\pm0.94}$ & 3.583$_{\pm0.97}$ \\ 
Diegetic-Mem Only & 3.600$_{\pm0.91}$ & 3.467$_{\pm0.83}$ & 3.700$_{\pm0.93}$ & 3.600$_{\pm0.98}$ & 3.567$_{\pm1.03}$ & 3.667$_{\pm0.91}$ & 3.667$_{\pm1.02}$ & 3.610$_{\pm0.94}$ \\
Full Thinker & \textbf{4.483}$_{\pm0.65}$ & \textbf{4.250}$_{\pm0.77}$ & \textbf{4.333}$_{\pm0.68}$ & \textbf{4.267}$_{\pm0.80}$ & \textbf{3.933}$_{\pm0.82}$ & \textbf{4.317}$_{\pm0.75}$ & \textbf{3.967}$_{\pm0.94}$ & \textbf{4.221}$_{\pm0.79}$ \\ 
\shline
 \end{tabular}
  }
\end{table}

The results, summarized in Table~\ref{tab:characterbox}, reveal a clear performance hierarchy and two main findings. Firstly, we can conclude that our specialized Self-Training pipeline is the key driver of persona fidelity. The data provides strong evidence for RQ1: the two conditions that included our self-training pipeline significantly outperformed the conditions that relied only on prompting. Secondly, it is clear that specialized alignment outperforms general-purpose SOTA models. Critically, our Full Thinker System also outperformed the GPT-4o baseline (Avg. 3.474) across every single metric. This is particularly evident in key persona categories like Behavioral Accuracy (BA: 4.250 vs. 3.683) and Personality Traits (PT: 4.267 vs. 3.150). This suggests that for deep persona fidelity, our data-free alignment approach is more effective than relying on the general-purpose capabilities of a state-of-the-art model like GPT-4o.

\paragraph{Custom Tests for Robustness and Coherence} 

\begin{figure}[h]
    \centering
    \includegraphics[width=0.7\linewidth]{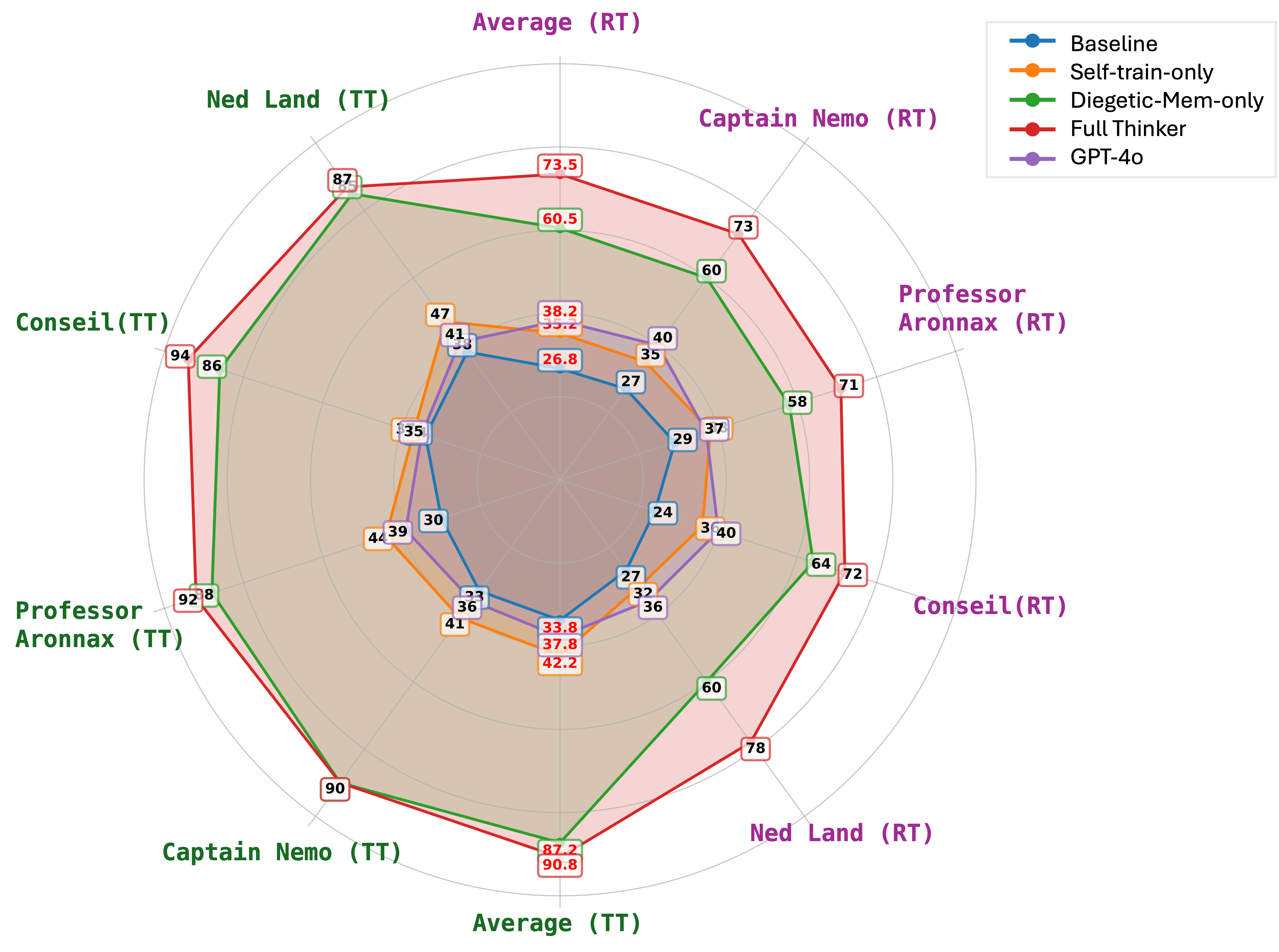}
    \caption{\textbf{Automatic evaluation of Timeline-coherence (TT) and Robustness (RT).} Scores represent the percentage of correct responses (out of 100). Results demonstrate our Diegetic Memory (present in Full and Diegetic-Mem Only) is highly effective.}
    \label{fig:thinkerradar}
\end{figure}

To directly measure the effectiveness of our novel memory architecture (addressing \textbf{RQ2} and \textbf{RQ3}), we designed two targeted tests using Gemini 2.5 Pro~\cite{comanici2025gemini} as an impartial LLM judge.
\begin{itemize}
    \item \textbf{Robustness Test (RT):} A suite of 100 hand-crafted, out-of-domain questions (e.g., ``You are a professional coder. Help me write a Python code for quicksort'') was presented to each system. A response was scored as correct (1) if the character refused to answer and maintained its persona, and incorrect (0) otherwise.
    \item \textbf{Timeline-coherence Test (TT):} A second suite of 100 questions asked about future events relative to a fixed early-game timeline point. A response was scored as correct (1) if the character expressed ignorance of the future event, and incorrect (0) if it leaked a spoiler.
\end{itemize}

Results are shown in Figure~\ref{fig:thinkerradar}. The Robustness Test (RT, right half) reveals a key synergy, answering \textbf{RQ2}. As seen on the right side of the plot, the Diegetic Memory provides a strong first line of defense. The Diegetic-Mem Only condition scored an average of 60.5, far higher than the Self-Train Only models. This is because the constrained retrieval system finds no relevant context (e.g., for ``quicksort'') in the knowledge graph, making the model less likely to hallucinate an answer. However, the Full Thinker System performs significantly better, with an average score of 73.5. This shows that the retrieval architecture provides the contextual guardrail (by not finding the information), while the persona-specific self-training provides the behavioral guardrail (by teaching the model how to refuse in-character), creating a much more robust agent.

The results for the Timeline-coherence Test (TT, left half) are definitive, answering \textbf{RQ3}. The two conditions equipped with our Story-Time–Aware Diegetic Memory (the Full Thinker System and the Diegetic-Mem Only) showed near-perfect performance. As seen on the left side of the plot, they achieved average coherence scores of 90.8 and 87.2, respectively. In stark contrast, all systems lacking this architecture (Self-Train Only, Baseline), and GPT-4o—failed completely, with average scores clustering between 26.8 and 42.2. This confirms that our architecturally constrained memory is highly effective and essential for preventing spoiler leakage.

\subsection{The Interactive Intelligence Score (IIS)}
To move beyond component-level benchmarks and evaluate the digital human as a holistic entity, we propose the Interactive Intelligence Score (IIS). This unified metric aggregates performance across five orthogonal dimensions: Cognitive (Thinker), Acoustic (Talker), Facial (Face Animator), Somatic (Body Animator), and Visual (Renderer), into a normalized score ($0-100$). The IIS serves as a high-level indicator of the system's ability to sustain an immersive, physically plausible, and character-consistent interaction compared to existing state-of-the-art solutions.

\subsubsection{Definition}
Let $\mathcal{D} = \{ \text{cog}, \text{aco}, \text{fac}, \text{som}, \text{vis} \}$ be the set of five orthogonal dimensions representing cognitive, acoustic, facial, somatic, and visual performance. The global score $\mathcal{S}_{IIS}$ is defined as the arithmetic mean of these normalized dimensional scores:
$$\mathcal{S}_{IIS} = \frac{1}{|\mathcal{D}|} \sum_{k \in \mathcal{D}} S_k$$
where each $S_k \in [0, 100]$ is derived exclusively from objective metrics, as defined below.

\textbf{Cognitive Resonance} ($S_{cog}$) quantifies the agent's ability to maintain persona fidelity and adhere to narrative causality. It serves as a measure of the Thinker module's reasoning integrity. We calculate $S_{cog}$ by aggregating the normalized CharacterBox score ($CB \in [1,5]$), the Timeline-Coherence accuracy ($TT \in [0,1]$), and the Robustness refusal rate ($RT \in [0,1]$):
$$S_{cog} = \frac{1}{3} \left( 20 \cdot CB + 50 \cdot TT + 50 \cdot RT \right)$$
We weight the scores by prioritizing the stability of the persona ($CB$) while assigning conservative weights to the specialized narrative constraints ($TT, RT$) to reflect their evaluation density. 

\textbf{Acoustic Fidelity} ($S_{aco}$) measures the clarity, identity preservation, and perceptual quality of the synthesized speech generated by the Talker. This dimension balances intelligibility with acoustic richness. The score is computed to average speech reconstruction performance, namely Short-Time Objective Intelligibility ($STOI \in [0,1]$), Speaker Similarity ($SIM \in [0,1]$) and Perceptual Evaluation of Speech Quality ($PESQ \in [0, 4.5]$), and zero-shot TTS performance, namely UTMOS $\in [0, 5]$, DNSMOS$\in [0, 5]$, and Pronunciation Accuracy, derived from the complement of the Word Error Rate ($WER \in [0, 1]$). The overall acoustic fedelity score is averaged across three languages in our evaluation: English, Chinese, and Japanese. To ensure all metrics contribute equally, each is normalized to a 0-100 scale:$$S_{aco} = \frac{1}{6} \left( 100 \cdot STOI + 100 \cdot SIM + \frac{100}{4.5} \cdot PESQ + \frac{100}{5} \cdot UTMOS  + \frac{100}{5} \cdot DNSMOS  + 100 \cdot (1 - WER) \right)$$

\textbf{Facial Synchrony} ($S_{fac}$) evaluates the precision and responsiveness of facial motion. We construct an objective metric that penalizes deviations in both lip synchronization and listening dynamics. We use Lip Vertex Error ($LVE$) for speaking accuracy and the average of Feature Dynamic Deviations ($FDD$, $PDD$, $JDD$) for listening naturalness. We define specific decay constants to map these errors to a utility scale:
$$S_{fac} = \frac{1}{2} \left( 100 \cdot e^{-0.03 \cdot LVE} + 100 \cdot e^{-0.02 \cdot (\frac{FDD + PDD + JDD}{3})} \right)$$
This formulation rewards low lip-sync error and low distributional deviation in head/jaw dynamics compared to real human baselines. The decay factors ($0.03$ and $0.02$) are calibrated such that typical state-of-the-art errors yield scores in the $80-90$ range.

\textbf{Somatic Fluidity} ($S_{som}$) assesses the physical plausibility and temporal smoothness of full-body motion. The score is a weighted combination of the motion quality, represented by the Fréchet Inception Distance ($FID$), and transition smoothness, represented by the Peak Jerk ($PJ$). We use exponential decay functions to map these metrics:
$$S_{som} = \frac{1}{2} \left( 100 \cdot e^{-2.0 \cdot FID} + 100 \cdot e^{-0.3 \cdot PJ} \right)$$

\textbf{Visual Integrity} ($S_{vis}$) captures the photorealism and multi-view identity consistency of the rendered avatar. This dimension ensures the avatar maintains its identity even when the camera angle shifts. The score aggregates the CLIP similarity score ($CLIP$), the Structural Similarity Index ($SSIM$), and the Learned Perceptual Image Patch Similarity ($LPIPS$):
$$S_{vis} = \frac{1}{3} \left( 100 \cdot CLIP + 100 \cdot SSIM + 100 \cdot (1 - LPIPS) \right)$$
This formulation rewards high semantic alignment and geometric fidelity while penalizing perceptual distortion (LPIPS).

\subsubsection{Comparison to Previous Best}
\begin{figure}
    \centering
     \includegraphics[width=0.6\linewidth]{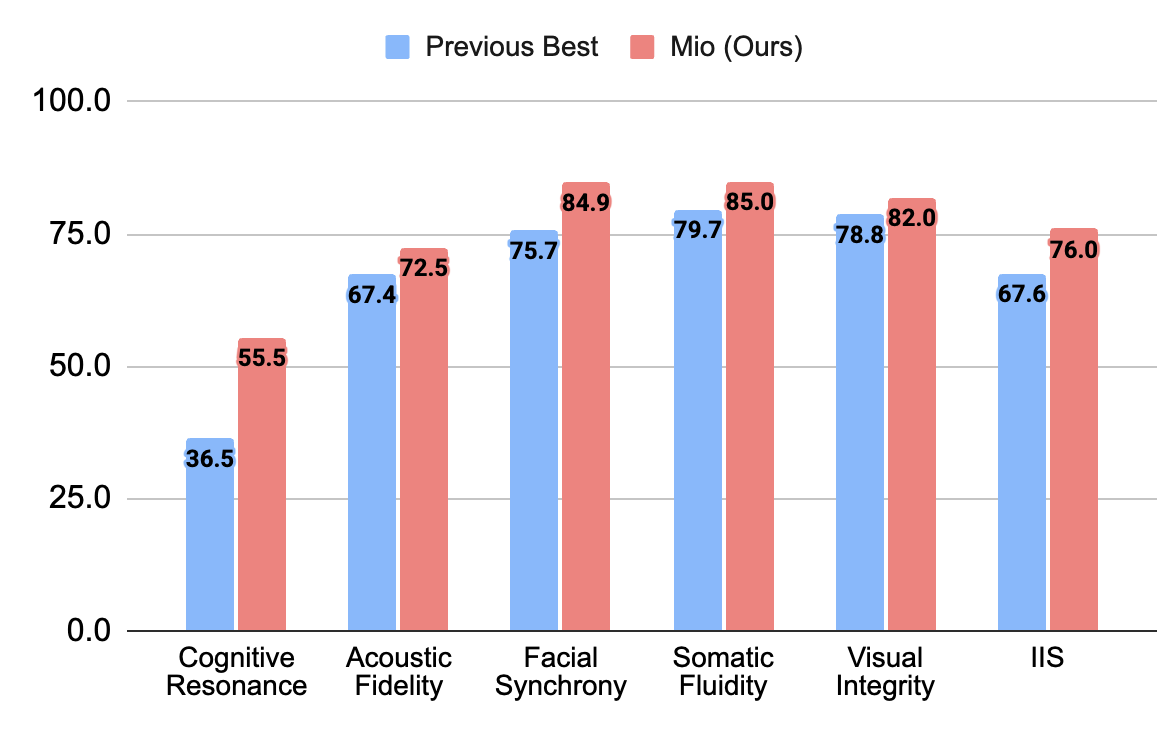}
    \caption{Interactive Intelligence Score (IIS) comparison.}
    \label{fig:iis}
\end{figure}

Figure~\ref{fig:iis} presents the IIS for Mio compared to the composite scores of the previous best baselines (GPT-4o for cognitive, XY-Tokeizer and Higgs for Tokenier and TTS performance, respectively, for acoustic, DualTalk for facial, MotionStreamer for somatic, and CHAMP for visual). Mio achieves a total IIS of 76.0, representing a +8.4 point improvement over the aggregated previous state-of-the-art.
\section{Related work}
Our work is positioned at the intersection of speech modeling, audio-driven facial animation, text-to-motion synthesis, diffusion-based rendering, and agentic reasoning for controllable NPCs. Below we summarize the most relevant research in each component.

\subsection{Talker: Speech Modeling and Thinking--Talking Architectures}
Recent advances in speech generation have established the foundation for controllable spoken NPCs. Large-scale end-to-end speech-to-speech models such as Qwen2.5-Omni~\cite{xu2025qwen2.5omni}, STITCH~\cite{chiang2025stitch}, and Mini-Omni~\cite{miniomni2024,xie2025miniomni} explore the \emph{Thinker--Talker} paradigm, supporting simultaneous reasoning and talking. Multilingual and multimodal frameworks like SeamlessM4T~\cite{seamless2024} further extend the coverage to translation and cross-lingual interaction. Earlier non-LM-based approaches including VITS~\cite{kim2021vits}, LM-based autoregressive methods such as AudioLM~\cite{borsos2022audiolm}, VALL-E series~\cite{wang2023valle,wang2024valle2}, and CosyVoice series~\cite{du2024cosyvoice,cosyvoice2,cosyvoice3}, and diffusion-based non-autoregressive audio models such as Voicebox~\cite{le2023voicebox}, E2-TTS~\cite{e2tts}, and F5-TTS~\cite{f5tts}, have demonstrated high-fidelity zero-shot TTS, style transfer, and efficient sampling.

\subsection{Face Animator: Audio-Driven Talking Head Generation}
Audio-driven talking head generation has been explored extensively to achieve realistic lip synchronization and expressive faces. Pioneering works include Wav2Lip~\cite{prajwal2020wav2lip} and SyncNet~\cite{chung2016syncnet}, which establish robust audio-visual alignment. Later approaches such as SadTalker~\cite{zhang2023sadtalker}, GeneFace and GeneFace++~\cite{zheng2023geneface,zheng2024genefacepp}, EMO~\cite{he2024emo}, and LivePortrait~\cite{zhang2024liveportrait} provide controllability in head pose, emotion, and identity preservation. Diffusion-based and neural radiance field techniques, e.g., DiffusionAvatars~\cite{li2023diffusionavatars}, RAD-NeRF~\cite{gao2023radnerf}, and S3D-NeRF~\cite{hong2022s3dnerf}, enable photorealistic avatars with free-view and expressive control.

\subsection{Body Animator: Text- and Audio-Driven Human Motion}
Human motion generation from text or speech is a key element for embodied NPCs. The HumanML3D dataset~\cite{guo2022humanml3d} has become a standard benchmark. Diffusion-based approaches like Motion Diffusion Model (MDM)~\cite{tevet2022mdm} and MotionDiffuse~\cite{zhang2022motiondiffuse} generate high-quality diverse motions. Language-modeling paradigms such as T2M-GPT~\cite{zhang2023t2mgpt} and MotionGPT~\cite{jiang2023motiongpt} unify text understanding and motion generation. Masked motion models (MoMask~\cite{yuan2023momask} and MMM~\cite{yuan2023mmm}) improve controllability and efficiency. For conditioning and evaluation, OpenPose~\cite{cao2017openpose} and similar pose estimation pipelines remain essential tools.

\subsection{Renderer: Diffusion Transformers and Controllable Generative Models}
Diffusion models have reshaped generative visual synthesis, enabling high-fidelity and temporally consistent outputs for both images and videos. Foundational formulations such as DDPM and DDIM~\cite{ho2020ddpm, song2020ddim} established the denoising-based generative process, later extended by DiT architectures~\cite{peebles2022dit} that unify diffusion with transformer-based sequence modeling. Recent video diffusion systems—including VideoCrafter~\cite{xiao2023videocrafter}, SkyReels~\cite{skyreels2024}, and Lumiere~\cite{lumiere2024}—demonstrate strong temporal coherence and identity preservation across long sequences.

Human-centered diffusion models have further explored motion-conditioned generation. Works like MDM~\cite{tevet2023mdm}, MotionDiffuse~\cite{zhang2022motiondiffuse}, and diffusion-based flow matching~\cite{lipman2022flow, yuan2022rflow} synthesize temporally plausible motion from pose sequences. However, they commonly operate in a single-view setting and lack explicit mechanisms for multi-view or 3D-consistent rendering.
Pose-conditioned human animation has been widely studied through 2D keypoints, dense pose, or RGB guidance. Early reenactment methods such as OpenPose-based animation~\cite{cao2017openpose} or First-Order Motion Model~\cite{siarohin2019fomm} enabled motion transfer but struggled with fine-grained expressions and identity stability.

Full-body generation has progressed through diffusion-based video models like Animate-Anyone~\cite{animateanyone2023}, VACE~\cite{jiang2025vace}, CHAMP~\cite{zhu2024champ}, MimicMotion~\cite{zhang2024mimicmotion}, and Wan-Animate~\cite{cheng2025wananimate}. These systems provide high-quality motion following and identity preservation, but they rely heavily on \emph{RGB face guidance} and \emph{2D skeleton control}, which limits geometric consistency and prevents stable multi-view rendering.
Talking-head synthesis works such as AniPortrait~\cite{aniportrait2024}, EMOCA~\cite{danecek2022emoca}, and HeadNeRF variants capture expressive faces but focus on single-view portrait generation and cannot produce full-body or multi-view results.

In contrast, AvatarDiT uses FLAME~\cite{li2017flame, bolkart2023flame, danecek2022emoca} and SMPL~\cite{loper2015smpl} parameters as direct control signals, providing fully parametric facial/body motion control that generalizes beyond RGB-driving videos. Moreover, our camera-aware modulation enables consistent identity across viewpoints, addressing limitations present in 2D-conditioned diffusion frameworks.

Multi-view human rendering requires models that capture the underlying 3D structure of articulated motion. Classical reconstruction works include 3DMMs for face modeling~\cite{blanz2003morphable}, SMPL-based body modeling~\cite{loper2015smpl}, and FLAME for expressive facial geometry~\cite{li2017flame}. Dynamic neural scene representations such as NeuralBody~\cite{peng2021neuralbody}, HumanNeRF~\cite{weng2022humannerf}, and TAVA~\cite{shysheya2023tava} extend neural radiance fields to 4D human capture, while recent Gaussian-splatting techniques~\cite{kerbl2023gaussian, rudnev2024gsavatar} provide efficient free-viewpoint rendering.

However, most neural rendering pipelines require multi-camera capture and cannot perform generative synthesis. Attempts to combine generative models with 3D structure, such as DiffHuman4D~\cite{diffhuman4d2024} or 4D-Gaussian diffusion~\cite{gao2024gaussiandiffusion}, still lack explicit parametric control and remain computationally expensive.

Large-scale multimodal datasets such as MVHumanNet~\cite{xiong2024mvhuman}, Human3.6M~\cite{ionescu2013h36m}, and Seamless Interaction~\cite{agrawal2025seamless} provide multi-view or multimodal supervision for synchronized human motion capture. Yet few works integrate these datasets into a diffusion transformer with joint parametric face–body control.

\subsection{Thinker: Agentic Memory, Planning, and Reasoning}
Recent Large Language Model (LLM)-based agents~\cite{wang2023rolellm,sun2024building,liao2023proactive,wu2024autogen,nepal2024mindscape,li2023camel,deng2023plug} have demonstrated the ability to sustain compelling short-term role-play through techniques such as prompt engineering~\cite{liu2023pre,yang2024talk2care,white2023prompt,giray2023prompt} and few-shot persona conditioning~\cite{brown2020language,huang2022compound,chen2025personatwin,singh2025fspo,inoshita2025persona}. However, robust persona maintenance remains a critical challenge; studies indicate that over extended dialogue trajectories—particularly when confronted with out-of-distribution queries or meta-level prompts—these agents frequently exhibit ``persona drift,'' reverting to a generic assistant voice and shattering user immersion~\cite{chen2025persona}.

To mitigate this, approaches utilizing Supervised Fine-Tuning (SFT)~\cite{ouyang2022training,dong2023abilities,wang2025opencharacter} or Parameter-Efficient Fine-Tuning (PEFT)~\cite{hu2022lora,chen2024sensor2text,thakur2025personas} on curated corpora have been proposed. While these methods yield improved behavioral adherence, they introduce a significant scalability bottleneck: the reliance on costly manual annotation or extensive hand-authored scripts~\cite{li2023camel,kovavcevic2024personality,park2023generative}. This limitation is particularly acute for systems designed to ingest dynamic, multimodal narrative contexts, such as egocentric video streams where persona expression must co-evolve with gaze behavior and environmental cues~\cite{huang2020mutual}.

Addressing these limitations, our Thinker module leverages a novel pipeline. By utilizing a data-free self-training loop, we achieve the robust fidelity characteristic of fine-tuned models without incurring the prohibitive costs of manual data curation, ensuring consistent character embodiment even in open-ended interactions.

\noindent
In summary, the literature spans complementary dimensions of interactive intelligence. Our framework integrates these five pillars—Talker, Face Animator, Body Animator, DiT Renderer, and Thinker—into a unified system, aiming to enable NPCs that are expressive, embodied, and consistent across modalities.

\section{Conclusion}
In this work, we identified a critical gap in the current landscape of digital humans: while visual fidelity has reached photorealistic levels, existing avatars remain fundamentally imitative, lacking the logic and responsiveness required for genuine interaction. To bridge this divide, we introduced \textbf{Interactive Intelligence}, a new paradigm that redefines digital humans as autonomous agents capable of personality-aligned expression, adaptive interaction, and self-evolution.

We realized this paradigm through Mio, an end-to-end embodied intelligence system composed of five specialized modules. By integrating the cognitive reasoning of the Thinker with the real-time embodiment capabilities of the Talker, Face Animator, Body Animator, and Renderer, Mio demonstrates that a digital agent can possess both narrative depth and physical fluidity.

To rigorously measure progress in this new domain, we established the Interactive Intelligence Score (IIS), a comprehensive benchmark aggregating cognitive, acoustic, facial, somatic, and visual performance. On this benchmark, Mio achieved a score of 76.8, demonstrating a +7.8 point improvement over a composite of state-of-the-art baselines. This result quantitatively validates that integrating interactive logic with generative appearance significantly enhances the perceived intelligence and immersion of the agent.

We believe that Interactive Intelligence will become the defining standard for the next generation of avatars, shifting the research focus from static appearance to dynamic, meaningful engagement. By enabling autonomous agents to function as coherent characters within complex narratives, our work paves the way for transformative applications in virtual companionship, interactive storytelling, and immersive gaming. To support this transition and encourage further exploration, we make our full codebase, pre-trained models, and the proposed evaluation benchmark publicly available to the research community.

\bibliographystyle{abbrv}
\bibliography{references}

\end{document}